\documentclass{article}

\PassOptionsToPackage{numbers, compress}{natbib}

\usepackage[final]{neurips_2021}

\usepackage[utf8]{inputenc} 
\usepackage[T1]{fontenc}    
\usepackage{hyperref}       
\usepackage{url}            
\usepackage{booktabs}       
\usepackage{amsfonts}       
\usepackage{nicefrac}       
\usepackage{microtype}      
\usepackage{xcolor}         
\usepackage{amsmath}
\usepackage{graphicx}
\usepackage{amssymb}
\usepackage{color,xcolor}
\usepackage{framed}
\usepackage{footmisc}
\usepackage{microtype}
\usepackage{subfigure}
\usepackage{multirow}
\usepackage{caption}
\usepackage{comment}

\title{A Unified Game-Theoretic Interpretation of Adversarial Robustness}

\author{%
	Jie Ren$^{a*}$,
    Die Zhang$^{a*}$,
    Yisen Wang$^{b}$\thanks{Equal contribution} ,
    Lu Chen$^{a}$,
    Zhanpeng Zhou$^{a}$,
    \\
    \textbf{Yiting Chen$^{a}$,
    Xu Cheng$^{a}$,
    Xin Wang$^{a}$,
    Meng Zhou$^{a, c}$\thanks{This work was done when Meng Zhou was an undergraduate at Shanghai Jiao Tong University.} ,
    Jie Shi$^{d}$,
    Quanshi Zhang}$^{a}$\thanks{Quanshi Zhang is the corresponding author. \texttt{zqs1022@sjtu.edu.cn}. This work is supervised by Dr. Quanshi Zhang. He is with the John Hopcroft Center and the MoE Key Lab of Artificial Intelligence, AI Institute, at the Shanghai Jiao Tong University, China.}
    \vspace{5pt}\\
    $^{a}$ Shanghai Jiao Tong University\\
    $^{b}$Key Lab. of Machine Perception (MoE), School of EECS, Peking University\\
    $^{c}$ Carnegie Mellon University\\
    $^{d}$ Huawei technologies Inc.
}

\begin{document}
	
	\maketitle
	
	\begin{abstract}
		This paper provides a unified view to explain different adversarial attacks and defense methods, \emph{i.e.} the view of multi-order interactions between input variables of DNNs.
		Based on the multi-order interaction, we discover that adversarial attacks mainly affect high-order interactions to fool the DNN.
		Furthermore, we find that the robustness of adversarially trained DNNs comes from category-specific low-order interactions.
		Our findings provide a potential method to unify adversarial perturbations and robustness, which can explain the existing defense methods in a principle way.
		Besides, our findings also make a revision of previous inaccurate understanding of the shape bias of adversarially learned features.	
		Our code is available online at \url{https://github.com/Jie-Ren/A-Unified-Game-Theoretic-Interpretation-of-Adversarial-Robustness}.
	\end{abstract}

	\section{Introduction}
	\label{sec:introduction}
	
	Adversarial robustness of deep neural networks (DNNs) has received increasing attention in recent years.
	Related studies include adversarial defense and attacks~\cite{szegedy2013intriguing,goodfellow2014explaining}.
	In terms of defense, adversarial training is an effective and the most widely-used method~\cite{madry2018towards,zhang2019theoretically,wang2019dynamic,wang2020improving,wu2020adversarial}.
	In spite of their fast development, the essential mechanism of the adversarial robustness is still unclear.
	Thus, the understanding of adversarial attacks and defense is an emerging direction in recent years.
	\citet{ilyas2019adversarial} demonstrated adversarial examples could be attributed to the presence of non-robust yet discriminative features.
	Some methods~\cite{gilmer2018relationship,weng2018evaluating} explored the mathematical bound for the model robustness.
	\citet{zhang2019interpreting,tsipras2018robustness} found adversarial training helped DNNs learn a more interpretable (more shape-biased) representation.
	Besides the feature interpretability,
	\citet{tsipras2018robustness} further showed an inherent tension between the adversarial robustness and the generalization power.
	
	Unlike above perspectives for explanations,
	we aim to propose a unified view to explain the essential reason why and how adversarial examples emerge, as well as essential mechanisms of various adversarial defense methods.
	We rethink the adversarial robustness from the novel perspective of interactions between input variables of a DNN.
	It is because in an adversarial example, adversarial perturbations on different pixels do not attack the DNN independently.
	Instead, perturbation pixels usually interact with each other to form a specific pattern for attacking.
	Surprisingly, we find that such interactions can explain various aspects of adversarial robustness.
	Specifically, this study aims to answer the following three questions.
	
	$\bullet$ \textbf{How to disentangle feature representations that are sensitive to adversarial perturbations.}
	Based on the game-theoretic interactions, we aim to summarize the distinct property of feature representations, which are sensitive to adversarial perturbations, among overall feature representations.
	
	$\bullet$ \textbf{How to explain the effectiveness of the adversarial training.}
	The above summarized property of sensitive feature representations also provides a new perspective to explain the utility of the adversarial training, \emph{i.e.} why and how the adversarial training penalizes such sensitive feature representations.
	
	$\bullet$ \textbf{How to unify various adversarial defense methods in a single theoretic system.}
	Our research provides a unified understanding for the success of adversarial defense methods~\cite{DBLP:journals/corr/abs-1906-03499,devries2017improved,jere2020singular}.

	\begin{figure}[t]
		\centering
		\begin{minipage}[t]{0.67\linewidth}
			\centering
			\includegraphics[width=\linewidth]{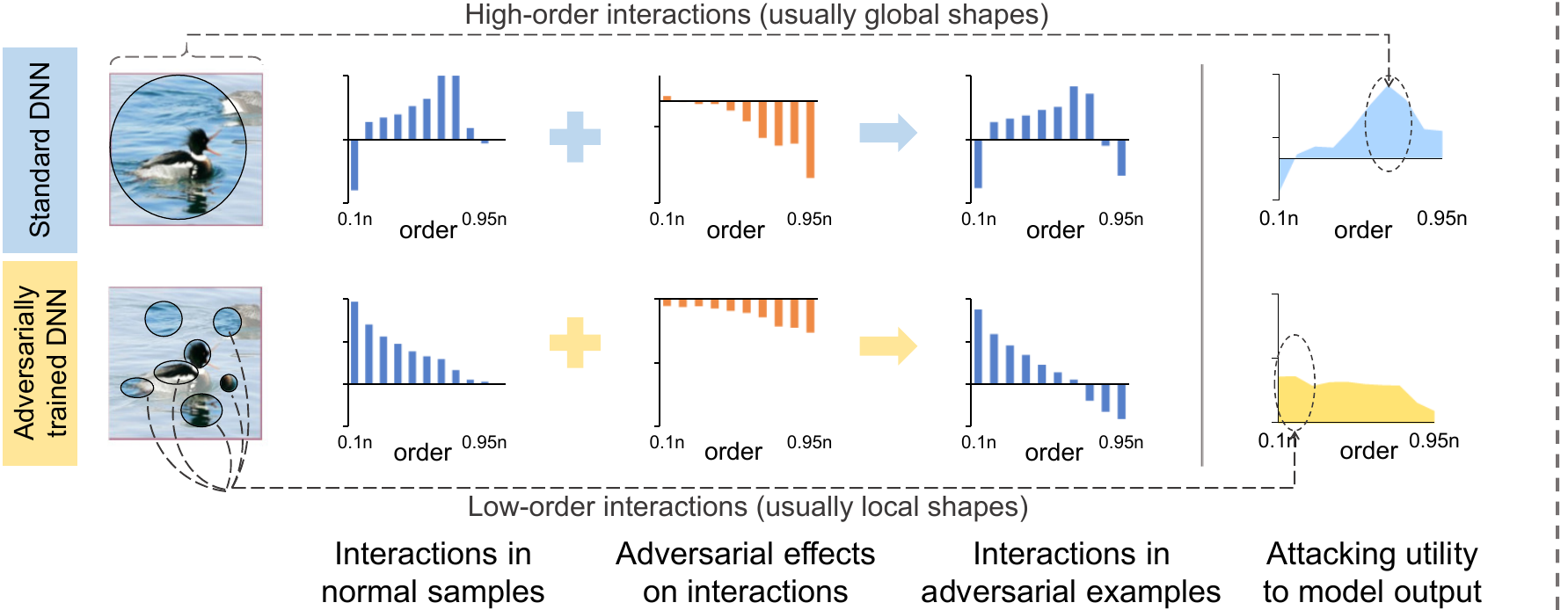}
		\end{minipage}
		\hfill
		\begin{minipage}[t]{0.28\linewidth}
			\centering
			\includegraphics[width=\linewidth]{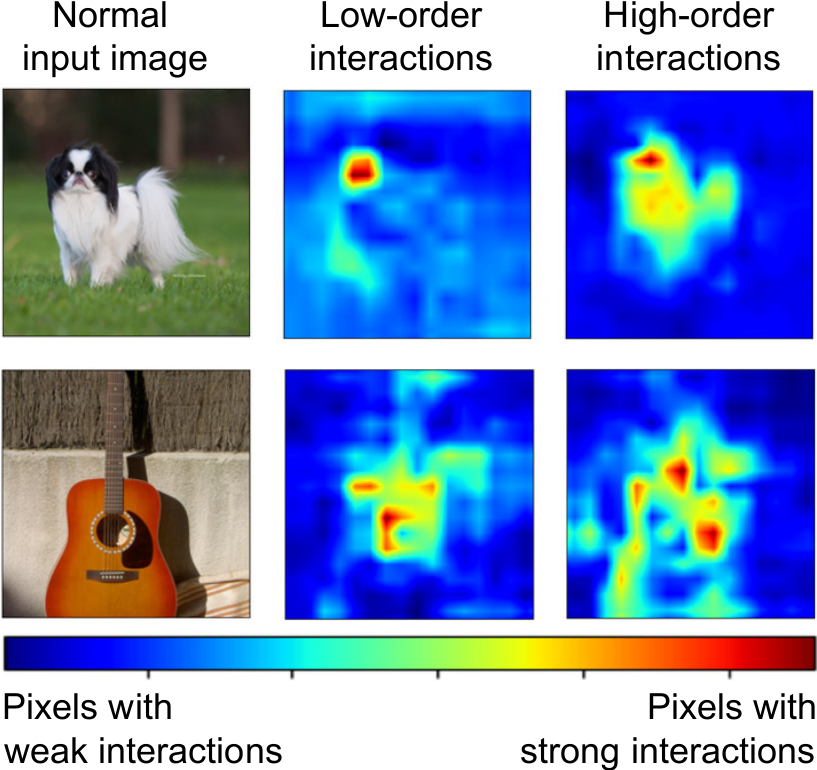}
		\end{minipage}
		\caption{(Left) Adversarial attacks mainly affect high-order interactions in input samples. High-order interactions in adversarially trained DNNs are more robust than those in standard DNNs. (Right) Regions with strong low-order and high-order interactions, which are visualized by the method extended from \cite{zhang2020interpreting-dropout}.}
		\label{fig:fig1}
		\vspace{-10pt}
	\end{figure}
	
	As a prerequisite of analyzing the adversarial robustness, let us first revisit the interaction between input variables in DNNs.
	Let a set of input variables collaborate with each other to form an inference pattern,
	the Shapley interaction index~\cite{grabisch1999axiomatic} is a standard metric to measure the numerical benefits to the inference from their collaborations.
	This metric can be further extended to the multi-order interaction~{\cite{zhang2020interpreting-dropout}}.
	For the interaction between two input variables ($i,j$), the interaction order measures the number of contextual variables that influence the significance of the interaction between $i$ and $j$.
	In other words, low-order interactions represent simple collaborations between input variables with small contexts, while high-order interactions indicate complex collaborations over large contexts (see Figure~\ref{fig:fig1} (right)).
	
	We further prove that the network output can be decomposed into the sum of multi-order interactions between different pairs of input variables.
	Thus, the overall effects of adversarial perturbations on the network output can be decomposed into elementary effects on different interaction components.
	Therefore, we can explain adversarial robustness using such elementary interaction components.

	(1)
	\textbf{We discover and partially prove that adversarial perturbations mainly affect high-order interactions, rather than low-order interactions} (see Figure~\ref{fig:fig1} (left)). In comparison, low-order interactions are naturally robust to attacks.
	In other words, adversarial attacks mainly affect the complex and large-scale collaborations among most pixels in the image.
	Based on this, we can successfully disentangle sensitive feature representations, \emph{i.e.} high-order interactions.

	\textit{Interaction vs. frequency \& rank.}
	Some studies explained adversarial examples as high-frequency features~\cite{yin2019fourier,wang2020high,harder2021spectraldefense} and high-rank features~\cite{jere2020singular}.
	We have conducted experiments to show that high-order interactions can better explain the essential property of attacking-sensitive representations, \emph{i.e.} the complex and large-scale visual concepts.

	(2)
	A clear difference between standard DNNs and adversarially trained DNNs is as follows.
	\textbf{Adversarial training significantly increases the robustness of high-order interactions.}
	In other words, attacks mainly affect complex collaborations in standard DNNs, while for adversarially trained DNNs, complex collaborations are not so vulnerable \emph{w.r.t.} simple collaborations.

	Then, we further explain the reason for the high robustness of adversarially trained DNNs.
	For interactions of each order, we define the disentanglement metric to identify whether interactions of this order are discriminative for the classification of a specific category, or represent common knowledge shared by different categories.
	For example, in Figure~\ref{fig:fig1} (left), interactions representing the blue water may be shared by different categories, while interactions corresponding to the head of the red-breasted merganser are discriminative for this category.
	Based on the disentanglement metric, we discover that \textbf{compared with standard DNNs, adversarially trained DNNs usually encode more discriminative low-order interactions.
		Discriminative low-order interactions make high-order interactions of adversarially trained DNNs robust to attacks,} because contexts of high-order interactions are composed of many small contexts of low-order interactions.
	For example, if the simple (low-order) interactions for the red-breasted merganser are learned to be discriminative, instead of being shared by the bicycle category, then it is difficult to attack this image towards the bicycle category.
	It is because it is difficult to use the low-order interactions of the red-breasted merganser's head to construct high-order interactions of bicycles.

	(3)
	\textbf{Our research provides a unified understanding for the success of several existing adversarial defense methods}, including the attribution-based detection of adversarial examples~\cite{DBLP:journals/corr/abs-1906-03499}, the recoverability of adversarial examples to normal samples, the cutout method~\cite{devries2017improved}, and
	the rank-based method~\cite{jere2020singular} (which is proved by \cite{jere2020singular} to be related to frequency-based methods~\cite{yin2019fourier,wang2020high,wu20143d}).

	\textbf{Above findings also slightly revise the previous explanation of adversarially learned features}~\cite{goodfellow2014explaining,tsipras2018robustness,dong2017towards,zhang2019interpreting}.
	They claimed that adversarial training learned more information about foreground shapes. We discover that these adversarially learned features are actually low-order interactions (usually local shapes), instead of modeling the global shape of the foreground.
	
    {\textbf{Explainable AI system based on game-theoretic interactions.}
	In fact, our research group led by Dr. Quanshi Zhang have proposed game-theoretic interactions, including interactions of different orders~\cite{zhang2020game} and multivariate interactions~\cite{zhang2021interpreting}.
	The interaction can be used as an typical metric to explain signal processing in DNNs from different perspectives.
	For example, {the game-theoretic metric} can be used to guide the learning of baseline values of Shapley values~\citep{ren2021learning}.
	Furthermore, we have built up a tree structure to explain the hierarchical interactions between words encoded by NLP models~\cite{zhang2021building}.
	We have also used interactions to explain the generalization power of DNNs~\cite{zhang2020interpreting-dropout}.
	The interaction can also explain how adversarial perturbations contribute to the attacking task~\cite{wang2021interpreting}, and explain the transferability of adversarial perturbations~\cite{wang2020unified}.
	Furthermore, we have also used the interaction to formulate the visual aesthetics~\cite{cheng2021hypothesis} and signal-processing properties of different types of visual concepts~\cite{cheng2021game} in DNNs.
	As an extension of the system of  game-theoretic  interactions,  in this  study,  we  explain  the  adversarial robustness  based  on interactions.}

	\section{Related work}
	\label{related_work}
	
	Attacking methods can be roughly summarized into
	white-box attacks~\cite{szegedy2013intriguing,goodfellow2014explaining,kurakin2016adversarial,papernot2016limitations,carlini2017towards} and black-box attacks~\cite{liu2016delving,papernot2017practical,chen2017zoo,bhagoji2018practical,ilyas2018black,wu2020skip,bai2020improving}.
	For defense, adversarial training is one of the most effective and widely-used defense methods~\cite{madry2018towards,zhang2019theoretically,wang2019dynamic,wang2020improving,wu2020adversarial}.
	Other defense methods include masking gradients~\cite{papernot2017practical,nayebi2017biologically}, modifying networks~\cite{cisse2017houdini,gao2017deepcloak}, and applying pre-processing on input images for testing~\cite{das2017keeping,meng2017magnet,xie2019feature,bai2019hilbert}.
	
	\textit{Explanations for adversarial examples.}
	Some previous studies focused on the reason for the existence of adversarial examples.
	\citet{goodfellow2014explaining} explained adversarial examples as a result of the high linearity of feature representations.
	\citet{gilmer2018relationship,ma2018characterizing} proved that the existence of adversarial examples was due to the geometry of the high-dimensional manifold.
	\citet{xie2019feature,xu2018structured,xu2019interpreting,bai2021improving} discovered that adversarial perturbations usually  activated substantial ``noise'' and semantically irrelevant features.
	\citet{engstrom2019exploring} investigated the vulnerability of DNNs to rotations and translations.
	\citet{tsipras2018robustness,ilyas2019adversarial} demonstrated that adversarial examples were attributed to non-robust yet discriminative features.

	\textit{Understandings of adversarial training.}
	\citet{athalye2018obfuscated} suggested that adversarial training did not cause the obfuscated gradients phenomenon, which boosted the robustness.
	\citet{goodfellow2014explaining,dong2017towards,tsipras2018robustness,zhang2019interpreting} found that adversarially trained DNNs learned more shape-biased  features than standard DNNs.
	\citet{chalasani2020concise} proved that adversarially trained DNNs \emph{w.r.t.} the $\ell_\infty$ attack exhibited more sparse attributions.
	\citet{tian2021analysis} showed that there is robustness imbalance among classes in adversarial training.
	\citet{song2018constructing} considered the adversarial training as the enumeration of all potential adversarial perturbations.
	\citet{wang2019dynamic} explained adversarial training from the perspective of min-max optimization.
	\citet{yin2019fourier,wang2018analyzing,harder2021spectraldefense} discovered adversarial training pushed DNNs to utilize low-frequency components in inputs.
	
	\textit{Understanding of the robustness.}
	\citet{szegedy2013intriguing,hein2017formal} computed Lipschitz constant to explain the robustness.
	\citet{ignatiev2019relating,boopathy2020proper} investigated the connection between network interpretability and adversarial robustness.
	\citet{tsipras2018robustness} proved the inherent tension between adversarial robustness and standard generalization power.
	\citet{fawzi2018analysis,gilmer2018relationship,weng2018evaluating} proved lower/upper bounds on the robustness.
	\citet{pal2020game} proposed a game-theoretic framework to understand the adversarial robustness, by formulating the game between attackers and defenders \emph{w.r.t.} a trained DNN. Under the game-theoretic framework, \citet{pal2020game} theoretically proved that the FGM attack and the random smoothing defense formed a Nash Equilibrium under some assumptions.

	Unlike previous explanations of the existence of adversarial examples and adversarial robustness, we explain adversarial examples and adversarial training from a new perspective, \emph{i.e.} the complexity of visual concepts that are learned by a DNN.

	\section{Decomposing attacking utility into interactions of multiple orders}
	\label{sec:interaction_shapley}
	
	\textbf{Preliminaries: using Shapley values~\cite{shapley1953value} and the Shapley interaction index~\cite{grabisch1999axiomatic} to explain a DNN.}
	\label{sec:shapley}
	The Shapley value~\cite{shapley1953value} in game theory is widely considered as an unbiased estimation for the importance or contribution of each player in a game.
	Given a trained DNN and the input with $n$ variables (\emph{e.g.} an image with $n$ pixels, a sentence with $n$ words)   $N=\{1,\cdots,n\}$, we can take input variables as players, and consider the network output as the reward.
	Shapley values can fairly divide and assign numerical effects on the network output to each input variable.
	More specifically, let   $2^N\overset{\text{def}}{=} \{S|S\subseteq N\}$ denote all potential subsets of $N$, and each subset $S$ represents a specific context (\emph{e.g.} a set of pixels in an image).
	$v(S)\in\mathbb{R}$ represents the scalar network output when  we keep variables in $S$ unchanged and mask variables in $N\setminus S$ by following settings in \cite{ancona2019explaining} (\emph{i.e.} setting a masked variable to the average value over different input samples).\footnote{For the DNN trained for multi-category classification, we set ${v(S)=\log p(y=c|}\text{given an input with variables in}~N\setminus S~\text{masked})$, where $c$ is selected as either the ground-truth category of the input, or the incorrectly classified category after attacks.}
	In particular, $v(N)$ refers to the network output \emph{w.r.t.} the entire input $N$ (\emph{e.g.} the whole image), and $v(\emptyset)$ denotes the output when we mask all variables.
	In this way, the Shapley value $\phi(i)$ unbiasedly measures the importance of the variable $i$ to the network output.
	\vspace{-5pt}
	\begin{equation}
		\begin{small}
			\phi(i) ={\sum}_{S\subseteq N\setminus \{i\}} p(S) \left[v(S\cup \{i\})-v(S)\right],\quad p(S)\overset{\text{def}}{=}\frac{(n-|S|-1)! |S|!}{n!}
		\end{small}
		\label{eq:shapley_value}
	\end{equation}
	In this way, the network output can be considered as the sum of elementary importances of input variables, \emph{i.e.} $v(N)=\sum_{i\in N}\phi(i)+v(\emptyset)$.
	The Shapley value has been proved to satisfy four desirable properties, \emph{i.e.} \textbf{\textit{linearity, nullity, symmetry}} and \textbf{\textit{efficiency}} properties, thereby being regarded as a fair method to allocate the network output to each input variable~\cite{weber1988probabilistic}.

	Input variables of a DNN do not contribute to the network output independently.
	Instead, different variables collaborate with each other to affect the network output, and we use the interaction to quantify the numerical utility of such collaborations.
	To this end, the Shapley interaction index~\cite{grabisch1999axiomatic} measures the utility of the collaboration between two input variables ($i,j$) by examining \textbf{whether the absence/presence of an input variable $j$ will change the importance of the other variable $i$.}
	Their interaction is defined as
	$I(i,j)=\tilde{\phi}(i)_{\text{$j$ always present}} - \tilde{\phi}(i)_{\text{$j$ always absent}}$,
	where $\tilde{\phi}(i)_{\text{$j$ always present}}$ and $\tilde{\phi}(i)_{\text{$j$ always absent}}$ denote the importance of the variable $i$ when $j$ is always present and when $j$ is always absent, respectively (please see \cite{grabisch1999axiomatic} for more details about $I(i,j)$).

	\begin{figure}[t]
		\centering
		\begin{minipage}[b]{0.67\linewidth}
			\includegraphics[width=\linewidth]{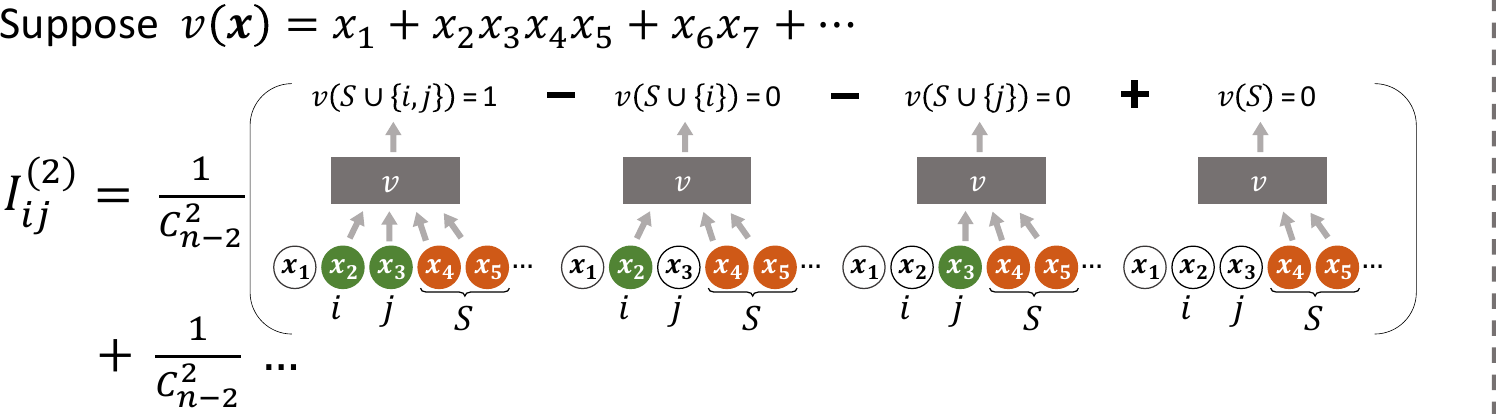}
		\end{minipage}
		\hfill
		\begin{minipage}[b]{0.32\linewidth}
			\includegraphics[width=\linewidth]{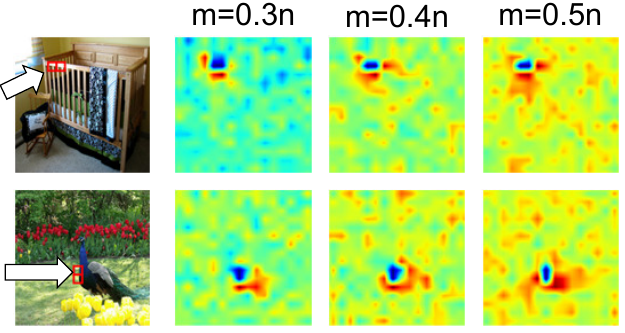}
		\end{minipage}
		\vspace{-15pt}
		\caption{(Left) Computation of the $m$-order interaction between two input variables ($m=2$). (Right) Contexts of the $m$-order interaction {\small $I^{(m)}_{ij}$}. Variables $(i,j)$ are indicated by red boxes. We extend the visualization method in \cite{zhang2021interpreting}, which visualized contexts of the Shapley interaction index, to show contexts of multi-order interactions (details in the supplementary material).
		Red (blue) colors indicate pixels that frequently (infrequently) appear in contexts which cause significant interactions.}
		\label{fig:interaction}
		\vspace{-5pt}
	\end{figure}

	\textbf{Decomposition of the existing Shapley interaction index into multiple orders.}
	{\citet{zhang2020interpreting-dropout}} further decomposes the Shapley interaction index $I(i,j)$~\cite{grabisch1999axiomatic} into interactions of different orders.
	\begin{equation}
		\begin{small}
			I(i,j) = \frac{1}{n-1}{\sum}_{m=0}^{n-2} I^{(m)}_{ij},\quad
			I^{(m)}_{ij} = {\mathbb{E}}_{S\subseteq N\setminus\{i,j\},|S|=m} \left[\Delta v(i,j,S) \right]
		\end{small}
	\end{equation}
	where {\small $\Delta v(i,j,S)= v(S\cup\{i,j\})-v(S\cup\{i\})-v(S\cup\{j\})+v(S)$}.
	
	\textbf{Proof of decomposing network output into multi-order interactions to explain adversarial attacks.}
	In this study, we successfully prove that the network output can be decomposed as the weighted sum of multi-order interactions.
	This enables us to use such interactions to explain adversarial attacks.
	Before the decomposition, let us first use interactions to understand the attack.
	As Figure~\ref{fig:interaction} (left) shows, the interaction component of each order {\small $I^{(m)}_{ij}$} represents collaborations between two input variables {\small $(i,j)$} with a specific contextual complexity.
	Let us consider input variables {\small $i,j$} and an arbitrary set of $m$ contextual variables.
	If {\small$I^{(m)}_{ij} > 0$}, it indicates that the presence of the variable $j$ will increase the importance of the variable $i$. Thus, we consider variables {\small $(i,j)$} have a positive interaction.
	If {\small$I^{(m)}_{ij}<0$}, it indicates a negative interaction.
	
	The $m$-order interaction between two input variables $(i,j)$ measures the average interaction between $(i,j)$ when we consider their collaborations with $m$ contextual variables ($m$ pixels).
	From another perspective, the order $m$ measures the number of contextual variables that influence the interaction between $i$ and $j$.
	For a low order $m$, {\small $I^{(m)}_{ij}$} reflects the interaction between $i$ and $j$ \emph{w.r.t.} simple contexts of a few variables.
	For a high order $m$, {\small $I^{(m)}_{ij}$} corresponds to the interaction \emph{w.r.t.} complex contexts with massive variables.
	Figure~\ref{fig:interaction} (right) visualizes the contexts $S$ corresponding to strong low-order interactions and strong high-order interactions, respectively.
	\citet{cheng2021game} has also proven that low-order interactions (local collaborations) mainly reflect simple and common concepts (features), and high-order interactions (global collaborations) usually represent complex and global features.

	\textit{$\bullet$ Properties of multi-order interactions.}
	{We have proven that {\small $I^{(m)}_{ij}$} satisfies \textbf{\textit{linearity, nullity, commutativity, symmetry}}, and \textbf{\textit{efficiency}}
	properties (please see the supplementary material for details).}
	In particular, the \textit{efficiency property} is given as follows.
	
	\textit{Efficiency property:}
	The output of the DNN given the entire input $v(N)$ can be decomposed into interactions
	of different orders, \emph{i.e.} {\small$v(N) = v(\emptyset)+\sum_{i\in N}\phi^{(0)}(i)+\sum_{i,j\in N,i\ne j}\sum_{m=0}^{n-2} J^{(m)}_{ij}$, where $J^{(m)}_{ij} \overset{\text{def}}{=} \frac{n-1-m}{n(n-1)} I^{(m)}_{ij}$, and $\phi^{(0)}(i) \overset{\text{def}}{=} v(\{i\})-v(\emptyset)$}.
	
	\textit{$\bullet$ Decomposing the overall attacking utility into interactions of different orders.}
	According to the  \textit{efficiency property} of interactions, the output of the DNN \emph{w.r.t.} the image $x$ can be decomposed into the sum of multi-order interactions between different pairs of input variables.
	Thus, we can decompose the overall utility of adversarial perturbations on the network output into elementary effects on interactions.
	\vspace{-5pt}
	\begin{equation}
		\begin{small}
			\Delta v(N|x)  \overset{\text{def}}{=} v(N|x)-v(N|x^{\text{adv}}) =  \underbrace{\Delta v(\emptyset|x) + \sum_{i\in N} \Delta \phi^{(0)}(i|x)}_{\text{usually can be ignored}} + \sum_{i\ne j\in N}\sum_{m=0}^{n-2} \Delta J^{(m)}_{ij}(x)
		\end{small}
		\label{eq:v(n)}
	\end{equation}
	where {\small $x\in \mathbb{R}^n$} denotes the normal sample, and {\small $x^{\text{adv}}=x+\Delta x \in \mathbb{R}^n$} denotes the adversarial example.
	{\small $v(N|x^{\text{adv}})$} denotes the network output given all variables in {\small $N$} of the adversarial example {\small $x^{\text{adv}}$}.
	{\small $v(N|x)$} corresponds to the normal sample.
	The first term {\small $\Delta v(\emptyset)=v(\emptyset|x)-v(\emptyset|x^{\text{adv}})=0$}.
	In the second term, {\small $\Delta \phi^{(0)}(i|x) \overset{\text{def}}{=} (v(\{i\}|x)-v(\emptyset|x))-(v(\{i\}|x^{\text{adv}})-v(\emptyset|x^{\text{adv}}))=v(\{i\}|x)-v(\{i\}|x^{\text{adv}})$}.
	Because in most applications, the importance of a single variable (\emph{e.g.} a pixel) is usually small, we can ignore this term.
	In the third term, {\small $\Delta J^{(m)}_{ij}(x)$} denotes the attacking utility of the $m$-order interaction between variables {\small $(i,j)$} in {\small $x$}.
	{\small$\Delta J^{(m)}_{ij}(x)\overset{\text{def}}{=}\frac{n-1-m}{n(n-1)} \Delta I^{(m)}_{ij}(x)$}, where {\small$\Delta I^{(m)}_{ij}(x)\overset{\text{def}}{=}I^{(m)}_{ij}(x)-I^{(m)}_{ij}(x^{\text{adv}})$} measures the elementary effects of adversarial perturbations on the $m$-order interaction.

	\section{Explaining adversarial attacks and defense using interactions}
	\label{sec:explain}

	To simplify the story, we only study the simplest and widely-used untargeted and targeted $\ell_\infty$ PGD attacks~\cite{madry2018towards}.
	
	\subsection{Attacks mainly affect high-order interactions}
	\label{sec:explain_attack}
	
	\textbf{Attacking utility of $m$-order interactions.}
	According to Eq.~\eqref{eq:v(n)}, effects of adversarial attacks mainly depend on changes of interactions {\small $\Delta I^{(m)}_{ij}(x)$} (or {\small $\Delta J^{(m)}_{ij}(x)$}).
	Thus, in this section, we further conduct experiments to show that adversarial attacks mainly affect {\small $\Delta I^{(m)}_{ij}(x)$} of high orders.
	In order to compare the multi-order interactions in normal samples and  those in adversarial examples, we measure the average interaction of a specific order $m$ among different pairs of variables in different input images, \emph{i.e.} {\small $I^{(m)}=\mathbb{E}_{x\in\Omega} \mathbb{E}_{i,j}[I^{(m)}_{ij}(x)]$},
	where {\small $\Omega\subseteq \mathbb{R}^n$} denotes the set of all samples.
	We use the normal samples {\small $\Omega_{\text{nor}}$} to compute {\small $I^{(m)}_{\text{nor}}$} and use adversarial examples {\small $\Omega_\text{adv}$} to compute {\small $I^{(m)}_{\text{adv}}$}.
	In this way, {\small $\Delta I^{(m)}\overset{\text{def}}{=}I^{(m)}_{\text{nor}}-I^{(m)}_{\text{adv}}$} represents effects of attacks on the $m$-order interactions.
	Similarly, {\small $\Delta J^{(m)}\overset{\text{def}}{=} \mathbb{E}_{x\in\Omega}\mathbb{E}_{i,j}[\Delta J^{(m)}_{ij}(x)]=\frac{n-1-m}{n(n-1)}\Delta I^{(m)}$} measures the attacking utility of the $m$-order interactions according to Eq.~\eqref{eq:v(n)}.

	\begin{figure}[t]
		\centering
		\begin{minipage}[b]{0.43\linewidth}
			\centering
			\includegraphics[width=\linewidth]{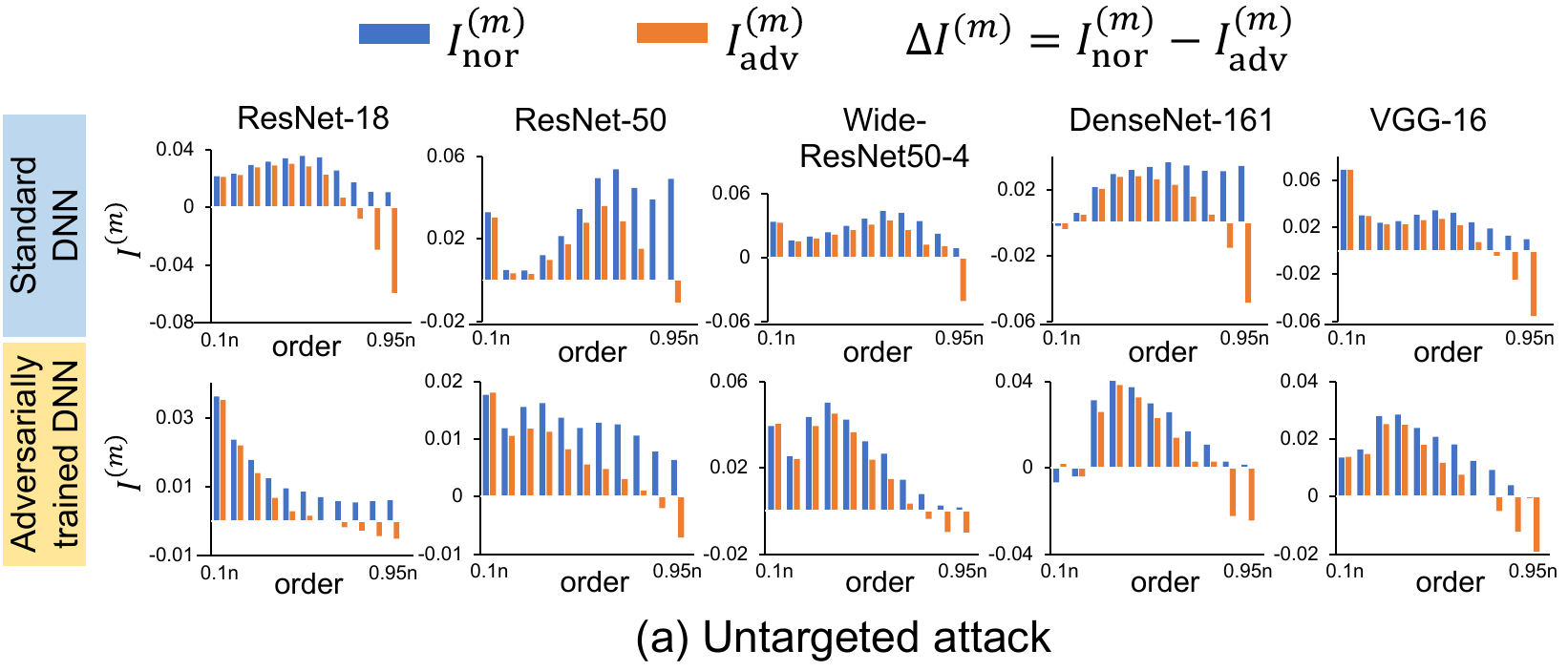}
		\end{minipage}
		\begin{minipage}[b]{0.56\linewidth}
			\centering
			\includegraphics[width=\linewidth]{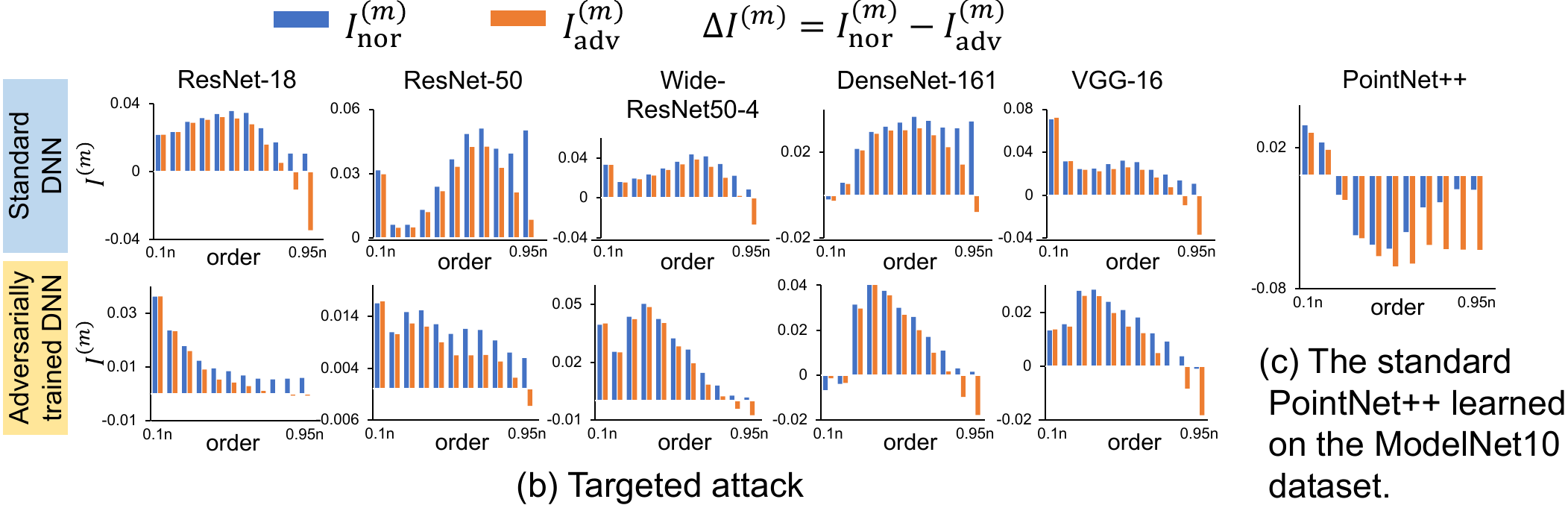}
		\end{minipage}
		\vspace{-15pt}
		\caption{The multi-order interaction $I^{(m)}_\text{nor}$ and $I^{(m)}_\text{adv}$ of standard DNNs and adversarially trained DNNs. Adversarial perturbations mainly affected high-order interactions.}
		\label{fig:explain_attack}
		\vspace{-10pt}
	\end{figure}
	
	\textbf{Comparing attacking utilities between interactions of different orders (in terms of {\small $I^{(m)}$}).}  In order to measure {\small $I^{(m)}_\text{nor}$} and {\small $I^{(m)}_\text{adv}$}, we conducted experiments on ResNet-18/34/50~\cite{he2016deep}, Wide-ResNet50-4~\citep{zagoruyko2016wide}, DenseNet-161~\cite{huang2017densely}, and VGG-16~\cite{simonyan2015very}.
	For each DNN, we obtained both the standardly trained version and  the adversarially trained version on the ImageNet dataset~\cite{imagenet2015}\footnote{We used pretrained models released by \citet{salman2020adversarially}.}.
	We also conducted experiments on PointNet++~\cite{qi2017pointnet} learned on the ModelNet10 dataset~\cite{wu20143d}, which is a 3D dataset.
	We measured the interaction {\small $I^{(m)}$} by setting {\small $v(S|x)=\log p(y=y^\text{truth}|\text{given variables in}~S~\text{in the input}~x~\text{and mask variables in}~N\setminus S)$}, and {\small $x\in \Omega$} was sampled from the validation set.
	By following settings in \cite{ancona2019explaining}, the masked variables in {\small $N\setminus S$} were set to the average value over different input samples.
	However, the computational cost of {\small $ I^{(m)}$} was intolerable.
	To reduce the computational cost, we applied the sampling-based approximation method in \cite{zhang2020interpreting-dropout} and
	computed interactions at the grid level, rather than the pixel level. Please see the supplementary material for more discussions.

	We considered both untargeted and targeted\footnote{In the untargeted attack, we considered the misclassified category as the target category. In the targeted attack, the target label was set as the misclassified category in the untarget attack.\label{fn:target} When we consider interaction \emph{w.r.t.} the target category, we set {\small $v(S) = \log p(y=y^{\text{target}}|\text{given variables in}~S~\text{and mask variables in}~N\setminus S)$}.} PGD attacks~\cite{madry2018towards} with the {\small $\ell_{\infty}$} constraint {\small $\Vert \Delta x \Vert_{\ell_\infty} \le \epsilon$} to generate adversarial examples. We set $\epsilon=32/255$ by following the setting in~\cite{xie2019feature}.
	The step size was set to $2/255$.
	For fair comparisons, we controlled the perturbation generated for each image to have similar attacking utility of 8.
	The attacking utility in untargeted attacks was defined as {\small $U_\text{untarget}(x) = h_{y^\text{truth}}(N|x)-h_{y^\text{truth}}(N|x^{\text{adv}})$}, where
	{\small $h_{y^\text{truth}}(N|x)$} and {\small $h_{y^\text{truth}}(N|x^{\text{adv}})$} denote network outputs of the ground-truth category {\small $y^\text{truth}$} before the softmax layer, when taking the normal sample and the adversarial example as input, respectively.
	The attacking utility in targeted attacks was defined as {\small $U_\text{target}(x) = (h_{y^\text{target}}(N|x^{\text{adv}})-h_{y^\text{truth}}(N|x^{\text{adv}}))-(h_{y^\text{target}}(N|x)-h_{y^\text{truth}}(N|x))$}, where {\small $h_{y^\text{target}}(\cdot)$} denote the network output of the target category\footref{fn:target} {\small $y^\text{target}$}.

	\textit{Results.} Figure~\ref{fig:explain_attack} shows the multi-order interactions within normal samples and adversarial examples.
	We found that adversarial perturbations significantly decreased high-order interactions, in both adversarially trained DNNs and standard DNNs.
	For example, in Figure~\ref{fig:explain_attack} (a), low-order interactions ($m<0.5 n$) in standard ResNet-18 decreased a little, while the interaction of order $m=0.95n$ dropped from $0.01$ to $-0.059$.
	This phenomenon was consistent with the heuristic findings of \citet{dong2017towards} that neurons corresponding to high-level semantics were ambiguous.

	\textit{Additional experiments on the target category and other categories.} Beyond above analysis about interactions \emph{w.r.t.} the ground-truth category, we also conducted experiments and found that high-order interactions \emph{w.r.t.} the target category\footref{fn:target} increased significantly.
	Please see the supplementary material for details.
	In sum, high-order interactions were much more sensitive than low-order interactions, which verified our conclusions.

	\textbf{Theoretic explanation of the sensitivity of high-order interactions.}
	Besides the above empirical observation, we can also theoretically prove that high-order interactions are more sensitive than low-order interactions.

	\textit{\textbf{Proposition 1} (equivalence between the multi-order interaction and the mutual information):
		Given an input sample $x\in X \subseteq \mathbb{R}^n$ and the network output $Y$, we define $X_S=\{x_S|x\in X\}$ where $S\subseteq N$; each $x_S$ represents the sample, where variables not in $S$ are masked.
		If $v(S)$ is set as the entropy of classification {\small$v(S) = H(Y|X_S)=\sum_{x_S} p(x_S) H(Y|X_S= x_S)$},} then {\small $I^{(m)}_{ij} =\mathbb{E}_{S\subseteq N\setminus\{i,j\},|S|=m} MI(X_i;X_j;Y|X_S)$}.
	The conditional mutual information $MI(X_i;X_j;Y|X_S)$ measures the remaining mutual information\footnote{Note that unlike the bivariate mutual information, $MI(X_i;X_j;Y|X_S)$ can be negative.} between $X_i,X_j$ and $Y$, when $X_S$ is given.
	When $X_S$ (with $m$ pixels) is given, $MI(X_i;X_j;Y|X_S)$ can be considered as the benefit to the inference from the interaction between $X_i$ and $X_j$.
	Please see the supplementary material for proofs.

	Proposition 1 indicates that compared to low-order interactions, high-order interactions are conditioned on larger contexts $S$, \emph{i.e.} conditioned on more contextual perturbations, thereby suffering more from adversarial perturbations.

	Note that in Proposition 1, we set {\small $\hat{v}(S)=H(Y|X_S)$}, which is slightly different from setting {\small $v(S)=\log p(y=y^\text{truth}|S,x)$} in Section~\ref{sec:explain_attack}. Nevertheless, the trend of {\small $v(S)$} can roughly reflect the negative trend of {\small $\hat{v}(S)$}, {which is discussed in the supplementary material}.

	\textbf{Comparison with frequency-based methods.}
	Previous studies~\cite{yin2019fourier,wang2020high,harder2021spectraldefense} explained adversarial perturbations as high-frequency features.
	Figure~\ref{fig:frequency} (left) shows the difference in the frequency between normal samples and adversarial samples.
	We find that high-order interactions can better distinguish adversarial examples and normal samples, than the frequency of features, which indicates that the interaction metric can better explain the essential property of adversarial perturbations.
	Please see the supplementary material for more details.

	\subsection{Explaining specific interactions encoded by adversarially trained DNNs}
	\label{sec:explain_training}
	
\begin{figure}[t]
		\centering
		\begin{minipage}[t]{0.48\linewidth}
			\centering
			\includegraphics[width=\linewidth]{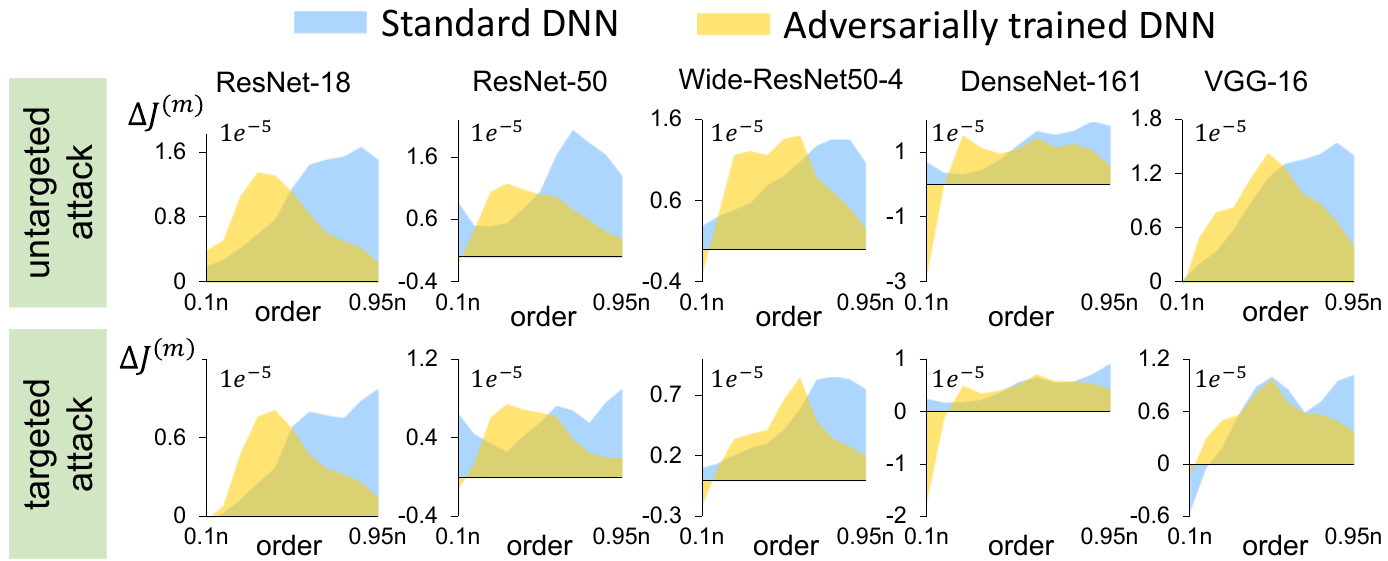}
			\vspace{-15pt}
			\caption{Distribution of compositional attacking utilities caused by interactions of different orders {\small $\{\Delta J^{(m)}\}$}.}
			\label{fig:diff}
		\end{minipage}
		\hfill
		\begin{minipage}[t]{0.48\linewidth}
			\centering
			\includegraphics[width=\linewidth]{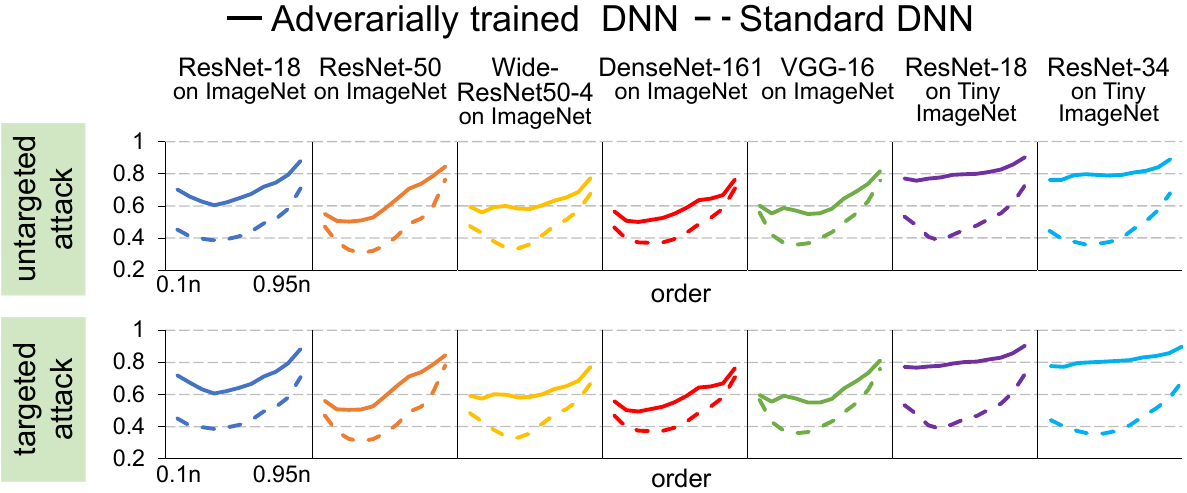}
			\vspace{-15pt}
			\caption{The interaction disentanglement {\small $D^{(m)}$}. Interactions of adversarially trained DNNs exhibited higher disentanglement than those of standard DNNs.}
			\label{fig:disentanglement}
		\end{minipage}
		\vspace{-5pt}
	\end{figure}

	\textbf{Discovery.} In order to understand the difference in signal-processing behaviors between adversarially trained DNNs and standard DNNs,
	we have proven that the overall attacking utility {\small $\mathbb{E}_{x\in\Omega} [\Delta v(N|x)]$} can be approximately decomposed as the sum of attacking utilities on multi-order interactions {\small $\Delta J^{(m)}$}, according to Eq.~\eqref{eq:v(n)}.
	Figure~\ref{fig:diff} shows the attacking utility of multi-order interactions \emph{i.e.} {\small $\Delta J^{(m)}=\frac{n-1-m}{n(n-1)} \Delta I^{(m)}$}.

	We noticed that for standard DNNs, adversarial attacks mainly affected high-order interactions.
	For adversarially trained DNNs, although high-order interactions were usually sensitive to attacks, low/middle-order interactions took up more attacking utilities.
	This phenomenon actually can also be observed in Figure~\ref{fig:explain_attack}, which is discussed in the supplementary material.
	Considering Eq.~\eqref{eq:v(n)}, this indicated that adversarial perturbations towards adversarially trained DNNs penalized both  complex features of global collaborations and simple features of local collaborations.

	\textbf{Explanation.} We further explain the above phenomena, \emph{i.e.} why high-order interactions in adversarially trained DNNs are more robust to attacks than those in standard DNNs.
	To this end, we propose the following disentanglement metric for the interaction of a specific order.
	\vspace{-3pt}
	\begin{equation}
		\begin{small}
			\begin{aligned}
				\!\!\!\! D^{(m)}\!\! =\! \mathbb{E}_{x\in \Omega}\mathbb{E}_{\substack{i,j\in N\\ i\ne j}}\frac{ |I^{(m)}_{ij}(x)|}{\mathbb{E}_{S\subseteq N\setminus\{i,j\}, |S|=m} |\Delta v(i,j,S|x)|}
				\!=\! \mathbb{E}_{x\in \Omega}\mathbb{E}_{\substack{i,j\in N\\ i\ne j}}\frac{ |\mathbb{E}_{S\subseteq N\setminus\{i,j\}, |S|=m} \Delta v(i,j,S|x)|}{ \mathbb{E}_{S\subseteq N\setminus\{i,j\}, |S|=m} |\Delta v(i,j,S|x)|}
			\end{aligned}
		\end{small}
	\end{equation}
	The above disentanglement metric examines whether or not the $m$-order interactions represent discriminative information for a specific category.
	The high value of $D^{(m)}$ indicates that the $m$-order interactions relatively purely describe specific categories. In other words, the interactions between $(i,j)$ under different contexts $S$ consistently have the same effects (either positive or negative) on the inference of a specific category.
	Let us consider the following toy example.
	When the pair of $(i,j)$ consistently have positive interactions towards a specific category under different contexts, \emph{i.e.} $\forall S\subseteq N\setminus\{i,j\},|S|=m, \Delta v(i,j,S|x) >0$, then we have $D^{(m)}=1$.
	This means that the $m$-th interactions between $i$ and $j$ stably promote the output probability of this category.
	In contrast, a low value of $D^{(m)}$ indicates that interactions between $(i,j)$ represent diverse categories.
	\emph{I.e.} given different contexts $S$, interactions between $(i,j)$ sometimes have positive effects on a specific category, and sometimes have negative effects.
	Please see the supplementary material for more discussions.
	
	\textit{Experiments.} Figure~\ref{fig:disentanglement} compares the interaction disentanglement $D^{(m)}$ of standard DNNs and the disentanglement of adversarially trained DNNs.
	Interactions of adversarially trained DNNs were more disentangled than those of standard DNNs, especially for low-order interactions.
	This indicated that low-order interactions in adversarially trained DNNs encoded more category-specific information for inference than low-order interactions in standard DNNs.

	Based on this observation, we could explain the robustness of high-order interactions to attacks in adversarially trained DNNs.
	Adversarial training learned more category-specific low-order interactions, which boosted the difficulty of attacking high-order interactions.
	This was because high-order interactions (usually global features) were usually constructed by low-order interactions (usually local features).
	Let us take the peacock in Figure~\ref{fig:interaction} (right) for example. In this image, low-order interactions (simple features) represented the body of the peacock and the green garden. In this case, it was difficult to attack this image to other categories (\emph{e.g.} the bicycle category) by constructing high-order interactions of the bicycle category using these peacock low-order features.
	Due to the difficulty of attacking high-order interactions, the adversarial examples towards the adversarially trained DNN had to attack low-order interactions, instead.

	\subsection{Unifying four existing adversarial defense methods}
	\label{sec:unifying}
	
	\textbf{$\bullet$ Explaining the attribution-based method of detecting adversarial examples.}
	\citet{DBLP:journals/corr/abs-1906-03499} proposed a method to use the attribution score of input variables to detect adversarial examples.
	In order to prove the effectiveness of this method, we define the multi-order Shapley value just like the multi-order interaction, as follows. {\small $\phi^{(m)}(i|x)\overset{\text{def}}{=} \mathbb{E}_{S\subseteq N\setminus\{i\},|S|=m}[v(S\cup \{i\}|x)-v(S|x)]$}, where $m$ denotes the order of the Shapley value.
	Then, we can prove that the attribution score which was used in~\cite{DBLP:journals/corr/abs-1906-03499} to detect adversarial exmaples, can be writtern as {\small $\phi^{(n-1)}(i|x)=v(N|x)-v({N\setminus\{i\}|x})$}.
	We further prove that $\phi^{(m)}(i|x)$ can be decomposed into interactions of different orders.
	\begin{equation}
		\begin{small}
			v(N|x)\! -\! v(\emptyset|x)\! =\! \frac{1}{n}\sum_{i\in N}\sum_{m=0}^{n-1}\phi^{(m)}(i|x),~~\phi^{(m)}(i|x) \!=\! \mathbb{E}_{j\in N\setminus\{i\}} \!\! \left[\sum_{k=0}^{m-1} I^{(k)}_{ij}(x)\right] \!\! + \phi^{(0)}(i|x)
			\label{eq:multi-order shapley}
		\end{small}
	\end{equation}
	According to the above equation, the overall adversarial effects can also be decomposed into elementary effects on $\phi^{(m)}(i|x)$.
	$\phi^{(0)}(i|x)=v(i|x)-v(\emptyset|x)$ is usually small and can be ignored, as discussed below Eq.~\eqref{eq:v(n)}.
	Among Shapley values of all orders, only the {\small$(n-1)$}-order component $\phi^{(n-1)}(i|x)$ contains the interactions of the highest order, which are the most sensitive interactions,
	according to Section~\ref{sec:explain_attack}.
	Therefore, the Shapley value component $\phi^{(n-1)}(i|x)$ is supposed to be the most sensitive to adversarial perturbations, which proves the effectiveness of the detection of adversarial examples.
	Please see the supplementary material for related proofs and discussions.

	\textbf{$\bullet$ Explaining high recoverability of adversarial examples towards adversarially trained DNNs.}
	The adversarial recoverability of a DNN is referred to as whether the DNN's adversarial examples can be inverted back to the normal sample by minimizing the classification loss.
	Specifically, given an adversarial example $x^{\text{adv}}$ generated by the untargeted attack~\cite{madry2018towards} via $\max_{\Vert x^{\text{adv}}-x \Vert _p\le \epsilon} \ell(x^{\text{adv}},y^{\text{truth}})$, we conduct a targeted attack~\cite{madry2018towards} on the adversarial sample to invert the classification result back to its ground-truth label, \emph{i.e.} $\min_{\delta^{\prime}}\ell(x^{\text{adv}}+\delta^{\prime},y^{\text{truth}})$. Let $\hat{x} = x^{\text{adv}}+\delta^{\prime}$  denote the recovered sample.
	If $\Vert \hat{x}-x \Vert_2 \le \Vert x^{\text{adv}}-x \Vert_2$, it indicates a high recoverability; otherwise, a low recoverability.

	We find that \textbf{\textit{adversarial examples generated by adversarially trained DNNs usually exhibit higher recoverability than those generated by standard DNNs.}}
	This phenomenon can be explained by Proposition 1 in Section~\ref{sec:explain_attack},
	\emph{i.e.} low-order interactions correspond to the mutual information conditioned on smaller adversarial perturbations, thereby suffering less from attacks.
	Therefore, we can partially explain the high recoverability of adversarial examples towards adversarially trained DNNs.
	It is because adversarially trained DNNs mainly focus on low-order interactions (see Section~\ref{sec:explain_training}), which leads to high recoverability.
	
	Besides, we conducted experiments on ResNet-18/50 and DenseNet-161 trained on the ImageNet dataset, which also validated the high recoverability of adversarial examples generated on adversarially trained DNNs.
	Please see the supplementary material for details.

	\begin{figure}[t]
		\centering
		\begin{minipage}[b]{0.49\linewidth}
			\centering
			\includegraphics[width=\linewidth]{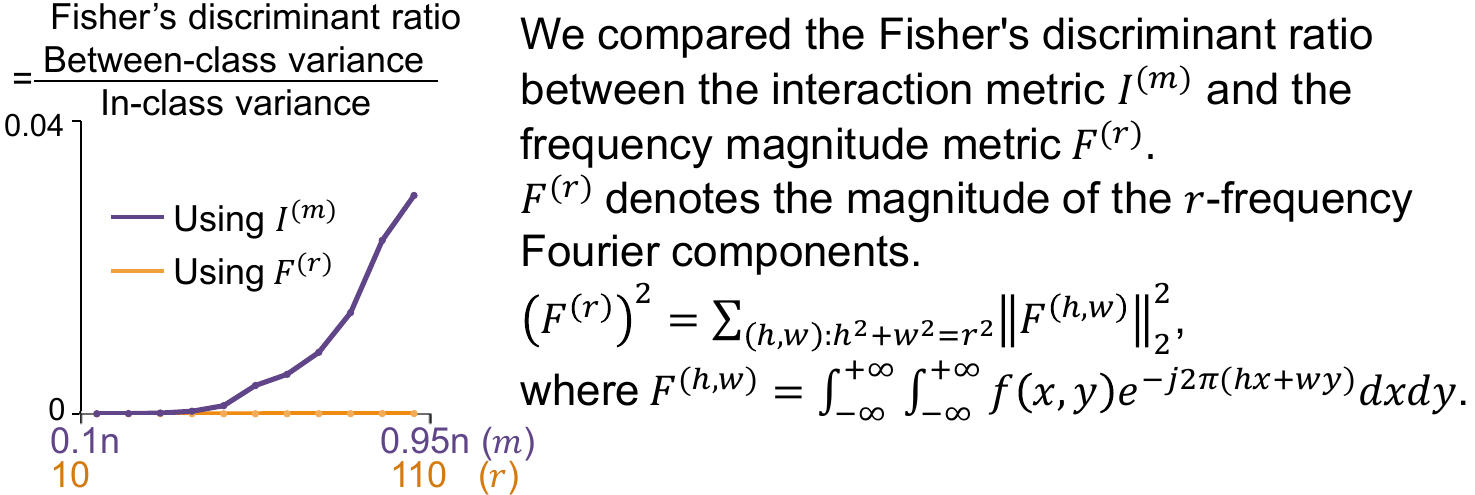}
		\end{minipage}
		\hfill
		\begin{minipage}[b]{0.49\linewidth}
			\centering
			\includegraphics[width=\linewidth]{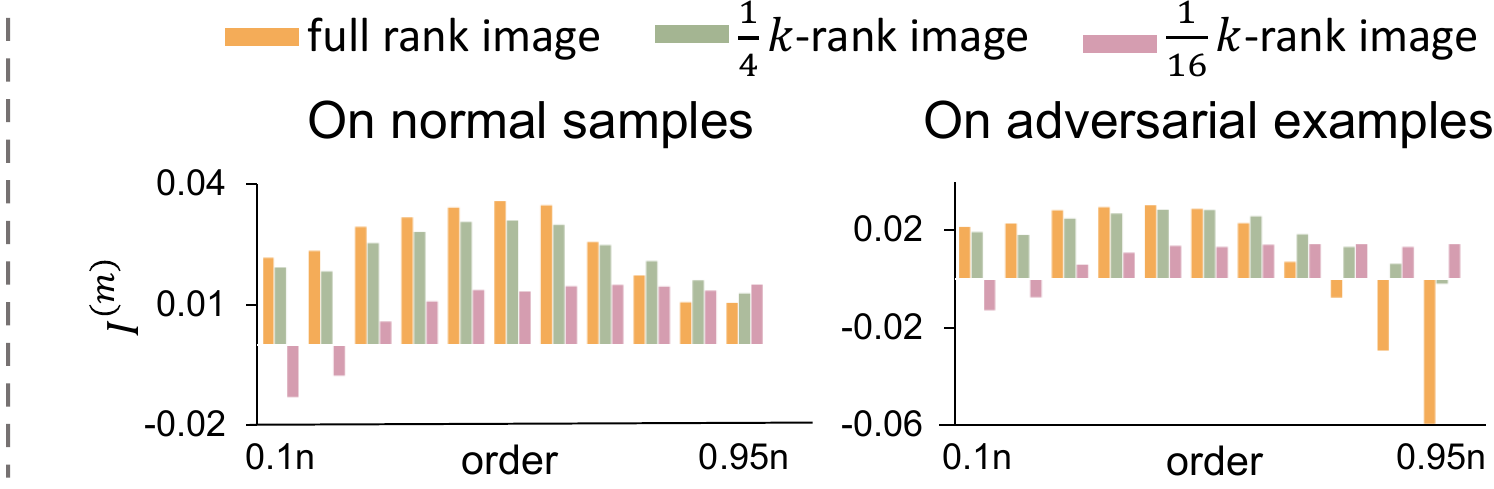}
		\end{minipage}
		\vspace{-5pt}
		\caption{(Left) The game-theoretic interaction is a more significant metric to distinguish normal samples and adversarial examples. (Right) Multi-order interactions in images of different ranks.}
		\vspace{-8pt}
		\label{fig:frequency}
	\end{figure}

	\textbf{$\bullet$ Explaining the rank-based method.}
	\citet{jere2020singular} discovered that standard DNNs paid certain attention to high-rank (\emph{w.r.t.} the Singular Value Decomposition) features, but adversarial training put more attention on low-rank features.
	This implied that adversarial examples mainly contained high-rank features.

	We used \cite{jere2020singular} to reduce the SVD rank of an image (either a normal sample or an adversarial example) to {\small $\frac{1}{4}k$}-rank image and {\small $\frac{1}{16}k$}-rank image, where $k$ denoted the full rank of the image.
	Figure~\ref{fig:frequency} (right) shows that
	(1) compared with normal samples, the rank-reducing operation on adversarial examples significantly increased high-order interactions.
	This indicated our conclusions in Section~\ref{sec:explain_attack} could explain that adversarial perturbations were mainly high-rank components in images.
	(2) When the rank of images was further reduced from {\small $\frac{1}{4}k$} to {\small $\frac{1}{16}k$}, the rank-reducing operation affected both low-order and high-order interactions. However, when we reduce the rank of either normal samples or adversarial examples, they exhibited similar effects on low-order interactions, but exhibited dramatically different effects on high-order interactions.
	This indicated that \textbf{the most discriminative factor of distinguishing adversarial examples from normal samples was high-order interactions, rather than the rank-reducing operation.}
	In other words, the game-theoretic interaction presents a more essential property of adversarial attacks.

	\begin{table}[t]
		\caption{Ratio of adversarial examples whose classification results were corrected.
			\label{tab:dropout}}
		\centering
		\resizebox{0.46\linewidth}{!}{
			\begin{tabular}{c|c|c|c|c}
				\hline
				\multirow{2}*{Standard DNNs} & \multirow{2}*{cutout} & \multicolumn{3}{|c}{ours} \\
				\cline{3-5}
				{} & {} & $\alpha=0.1$ & $\alpha=0.2$ & $\alpha=0.3$  \\
				\hline
				ResNet-18 & 7.12\% & 19.62\% & 35.55\% & \textbf{44.14\%} \\
				ResNet-50 & 10.63\% & 28.53\% & 45.53\% & \textbf{52.33\%} \\
				DenseNet-161 & 11.32\% & 27.16\% & 46.15\% & \textbf{56.35\%} \\
				\hline
		\end{tabular}}
		\hspace{3pt}
		\resizebox{0.46\linewidth}{!}{
			\begin{tabular}{c|c|c|c|c}
				\hline
				{$~$ Adversarially$~~$}  & \multirow{2}*{cutout} & \multicolumn{3}{|c}{ours} \\
				\cline{3-5}
				{$~$ trained DNNs$~~$}  & {} & $\alpha=0.1$ & $\alpha=0.2$ & $\alpha=0.3$  \\
				\hline
				ResNet-18 & 23.92\% & 22.46\% & 33.78\% & \textbf{39.18\%} \\
				ResNet-50 & 24.27\% & 23.23\% & 35.23\% & \textbf{41.54\%} \\
				DenseNet-161 & 35.65\% &  24.65\% & 39.25\% & \textbf{45.74\%} \\
				\hline
		\end{tabular}}
		\vspace{-12pt}
	\end{table}

	\textbf{$\bullet$ Explaining the cutout method.}
	\citet{devries2017improved} proposed the cutout method to boost the robustness of DNNs by masking out a square region of the input during training.
	This operation can be roughly considered as applying a random dropout operation~\cite{srivastava2014dropout} to input images.
	The dropout operation on input images has been proven to destroy interactions between input variables~\cite{zhang2020interpreting-dropout}.
	To this end, we have further proven that the dropout operation mainly hurts high-order interactions, rather than low-order interactions (please see the proof in the supplementary material).
	In this way, the cutout method can be considered as removing high-order interactions and pushing DNNs to learn more insensitive low-order interactions.
	Therefore, the learned DNN is robust to adversarial attacks.
	
	In experiments, we applied the dropout operation to the misclassified adversarial examples, in order to verify the attacking-alleviating influence on the network output of removing high-order interactions.
	If the removal of high-order interactions could alleviate the effects of adversarial attacks, it would prove the attacking-alleviating influence of removing high-order interactions.
	Given ResNet-18/50 and DenseNet-161 trained on the ImageNet dataset,
	we selected normal samples that are correctly classified in the validation set of the ImageNet dataset, and generated adversarial examples with the $\ell_\infty$ untargeted PGD attack based on~\cite{xie2019feature}.
	The attack was stopped once it succeeded.
	We applied the dropout operation with different dropout rates $\alpha$ at the pixel level, and the dropped pixels were filled with the average value of surrounding pixels.
	For the cutout method, we set the length of the side of the masked square regions as $112$, which was half of the side length of the input sample (dropping $\alpha=25\%$ pixels of the input), following settings in \cite{devries2017improved}.
	Table~\ref{tab:dropout} reports the ratio of adversarial examples whose classification results were corrected.
	When we applied dropout to adversarial examples, the classification accuracy of adversarial examples significantly increased,
	and even performed better than the cutout method.
	This indicated that the removal of high-order interactions presented a more essential defense mechanism than the cutout method.

	\section{Conclusion}
	\label{sec:conclusion}
	
	In this paper, we have used the multi-order interaction to provide a unified understanding for the success of adversarial attacks and defense.
	Based on the multi-order interaction, we have explained adversarial attacking as mainly affecting high-order interactions between input variables.
	Furthermore, we have also explained the success of adversarial training, \emph{i.e.} learning more category-specific low-order interactions to boost the robustness.
	Besides, we have further provided a unified explanation for some existing adversarial defense methods.
	
	\begin{ack}
	{This work is partially supported by the National Nature Science Foundation of China (No. 61906120, U19B2043), Shanghai Natural Science Foundation (21JC1403800,21ZR1434600), Shanghai Municipal Science and Technology Major Project (2021SHZDZX0102).
	This work is also partially supported by Huawei Technologies Inc.
	Prof. Yisen Wang is partially supported by the National Natural Science Foundation of China under Grant 62006153, and Project 2020BD006 supported by PKU-Baidu Fund.
	Xin Wang is supported by Wu Wen Jun Honorary Doctoral Scholarship, AI Institute, Shanghai Jiao Tong University.}
	\end{ack}

	\newpage
	{
		\small
		\bibliographystyle{plainnat}
		\bibliography{interaction_adv}

\begin{thebibliography}{86}
\providecommand{\natexlab}[1]{#1}
\providecommand{\url}[1]{\texttt{#1}}
\expandafter\ifx\csname urlstyle\endcsname\relax
  \providecommand{\doi}[1]{doi: #1}\else
  \providecommand{\doi}{doi: \begingroup \urlstyle{rm}\Url}\fi

\bibitem[Ancona et~al.(2019)Ancona, Oztireli, and Gross]{ancona2019explaining}
Marco Ancona, Cengiz Oztireli, and Markus Gross.
\newblock Explaining deep neural networks with a polynomial time algorithm for
  shapley value approximation.
\newblock In \emph{ICML}, 2019.

\bibitem[Athalye et~al.(2018)Athalye, Carlini, and
  Wagner]{athalye2018obfuscated}
Anish Athalye, Nicholas Carlini, and David Wagner.
\newblock Obfuscated gradients give a false sense of security: Circumventing
  defenses to adversarial examples.
\newblock In \emph{ICML}, 2018.

\bibitem[Bai et~al.(2019)Bai, Feng, Wang, Dai, Xia, and Jiang]{bai2019hilbert}
Yang Bai, Yan Feng, Yisen Wang, Tao Dai, Shu-Tao Xia, and Yong Jiang.
\newblock Hilbert-based generative defense for adversarial examples.
\newblock In \emph{ICCV}, 2019.

\bibitem[Bai et~al.(2020)Bai, Zeng, Jiang, Wang, Xia, and
  Guo]{bai2020improving}
Yang Bai, Yuyuan Zeng, Yong Jiang, Yisen Wang, Shu-Tao Xia, and Weiwei Guo.
\newblock Improving query efficiency of black-box adversarial attack.
\newblock In \emph{ECCV}, 2020.

\bibitem[Bai et~al.(2021)Bai, Zeng, Jiang, Xia, Ma, and Wang]{bai2021improving}
Yang Bai, Yuyuan Zeng, Yong Jiang, Shu-Tao Xia, Xingjun Ma, and Yisen Wang.
\newblock Improving adversarial robustness via channel-wise activation
  suppressing.
\newblock In \emph{ICLR}, 2021.

\bibitem[Bhagoji et~al.(2018)Bhagoji, He, Li, and Song]{bhagoji2018practical}
Arjun~Nitin Bhagoji, Warren He, Bo~Li, and Dawn Song.
\newblock Practical black-box attacks on deep neural networks using efficient
  query mechanisms.
\newblock In \emph{ECCV}, 2018.

\bibitem[Boopathy et~al.(2020)Boopathy, Liu, Zhang, Liu, Chen, Chang, and
  Daniel]{boopathy2020proper}
Akhilan Boopathy, Sijia Liu, Gaoyuan Zhang, Cynthia Liu, Pin-Yu Chen, Shiyu
  Chang, and Luca Daniel.
\newblock Proper network interpretability helps adversarial robustness in
  classification.
\newblock In \emph{ICML}, 2020.

\bibitem[Carlini and Wagner(2017)]{carlini2017towards}
Nicholas Carlini and David Wagner.
\newblock Towards evaluating the robustness of neural networks.
\newblock In \emph{SP}, 2017.

\bibitem[Chalasani et~al.(2020)Chalasani, Chen, Chowdhury, Wu, and
  Jha]{chalasani2020concise}
Prasad Chalasani, Jiefeng Chen, Amrita~Roy Chowdhury, Xi~Wu, and Somesh Jha.
\newblock Concise explanations of neural networks using adversarial training.
\newblock In \emph{ICML}, 2020.

\bibitem[Chen et~al.(2017)Chen, Zhang, Sharma, Yi, and Hsieh]{chen2017zoo}
Pin-Yu Chen, Huan Zhang, Yash Sharma, Jinfeng Yi, and Cho-Jui Hsieh.
\newblock Zoo: Zeroth order optimization based black-box attacks to deep neural
  networks without training substitute models.
\newblock In \emph{Proceedings of the 10th ACM Workshop on Artificial
  Intelligence and Security}, 2017.

\bibitem[Cheng et~al.(2021{\natexlab{a}})Cheng, Chu, Zheng, Ren, and
  Zhang]{cheng2021game}
Xu~Cheng, Chuntung Chu, Yi~Zheng, Jie Ren, and Quanshi Zhang.
\newblock A game-theoretic taxonomy of visual concepts in dnns.
\newblock \emph{arXiv preprint arXiv:2106.10938}, 2021{\natexlab{a}}.

\bibitem[Cheng et~al.(2021{\natexlab{b}})Cheng, Wang, Xue, Liang, and
  Zhang]{cheng2021hypothesis}
Xu~Cheng, Xin Wang, Haotian Xue, Zhengyang Liang, and Quanshi Zhang.
\newblock A hypothesis for the aesthetic appreciation in neural networks.
\newblock \emph{arXiv preprint arXiv:2108.02646}, 2021{\natexlab{b}}.

\bibitem[Cisse et~al.(2017)Cisse, Adi, Neverova, and Keshet]{cisse2017houdini}
Moustapha Cisse, Yossi Adi, Natalia Neverova, and Joseph Keshet.
\newblock Houdini: Fooling deep structured prediction models.
\newblock \emph{arXiv preprint arXiv:1707.05373}, 2017.

\bibitem[Cohen et~al.(2019)Cohen, Rosenfeld, and Kolter]{cohen2019certified}
Jeremy Cohen, Elan Rosenfeld, and Zico Kolter.
\newblock Certified adversarial robustness via randomized smoothing.
\newblock In \emph{ICML}, 2019.

\bibitem[Covert et~al.(2020)Covert, Lundberg, and Lee]{covert2020understanding}
Ian Covert, Scott Lundberg, and Su-In Lee.
\newblock Understanding global feature contributions with additive importance
  measures.
\newblock \emph{NeurIPS}, 2020.

\bibitem[Cui et~al.(2019)Cui, Marttinen, and Kaski]{cui2019learning}
Tianyu Cui, Pekka Marttinen, and Samuel Kaski.
\newblock Learning global pairwise interactions with bayesian neural networks.
\newblock \emph{arXiv preprint arXiv:1901.08361}, 2019.

\bibitem[Das et~al.(2017)Das, Shanbhogue, Chen, Hohman, Chen, Kounavis, and
  Chau]{das2017keeping}
Nilaksh Das, Madhuri Shanbhogue, Shang-Tse Chen, Fred Hohman, Li~Chen,
  Michael~E Kounavis, and Duen~Horng Chau.
\newblock Keeping the bad guys out: Protecting and vaccinating deep learning
  with jpeg compression.
\newblock \emph{arXiv preprint arXiv:1705.02900}, 2017.

\bibitem[DeVries and Taylor(2017)]{devries2017improved}
Terrance DeVries and Graham~W Taylor.
\newblock Improved regularization of convolutional neural networks with cutout.
\newblock \emph{arXiv preprint arXiv:1708.04552}, 2017.

\bibitem[Dong et~al.(2017)Dong, Su, Zhu, and Bao]{dong2017towards}
Yinpeng Dong, Hang Su, Jun Zhu, and Fan Bao.
\newblock Towards interpretable deep neural networks by leveraging adversarial
  examples.
\newblock \emph{arXiv preprint arXiv:1708.05493}, 2017.

\bibitem[Engstrom et~al.(2019)Engstrom, Tran, Tsipras, Schmidt, and
  Madry]{engstrom2019exploring}
Logan Engstrom, Brandon Tran, Dimitris Tsipras, Ludwig Schmidt, and Aleksander
  Madry.
\newblock Exploring the landscape of spatial robustness.
\newblock In \emph{ICML}, 2019.

\bibitem[Fawzi et~al.(2018)Fawzi, Fawzi, and Frossard]{fawzi2018analysis}
Alhussein Fawzi, Omar Fawzi, and Pascal Frossard.
\newblock Analysis of classifiers’ robustness to adversarial perturbations.
\newblock \emph{Machine Learning}, 107\penalty0 (3):\penalty0 481--508, 2018.

\bibitem[Fisher(1936)]{fisher1936use}
Ronald~A Fisher.
\newblock The use of multiple measurements in taxonomic problems.
\newblock \emph{Annals of eugenics}, 7\penalty0 (2):\penalty0 179--188, 1936.

\bibitem[Gao et~al.(2017)Gao, Wang, Lin, Xu, and Qi]{gao2017deepcloak}
Ji~Gao, Beilun Wang, Zeming Lin, Weilin Xu, and Yanjun Qi.
\newblock Deepcloak: Masking deep neural network models for robustness against
  adversarial samples.
\newblock \emph{arXiv preprint arXiv:1702.06763}, 2017.

\bibitem[Gilmer et~al.(2018)Gilmer, Metz, Faghri, Schoenholz, Raghu,
  Wattenberg, Goodfellow, and Brain]{gilmer2018relationship}
Justin Gilmer, Luke Metz, Fartash Faghri, Samuel~S Schoenholz, Maithra Raghu,
  Martin Wattenberg, Ian Goodfellow, and G~Brain.
\newblock The relationship between high-dimensional geometry and adversarial
  examples.
\newblock \emph{arXiv preprint arXiv:1801.02774}, 2018.

\bibitem[Goodfellow et~al.(2014)Goodfellow, Shlens, and
  Szegedy]{goodfellow2014explaining}
Ian~J Goodfellow, Jonathon Shlens, and Christian Szegedy.
\newblock Explaining and harnessing adversarial examples.
\newblock \emph{arXiv preprint arXiv:1412.6572}, 2014.

\bibitem[Grabisch and Roubens(1999)]{grabisch1999axiomatic}
Michel Grabisch and Marc Roubens.
\newblock An axiomatic approach to the concept of interaction among players in
  cooperative games.
\newblock \emph{International Journal of Game Theory}, 28\penalty0
  (4):\penalty0 547--565, 1999.

\bibitem[Harder et~al.(2021)Harder, Pfreundt, Keuper, and
  Keuper]{harder2021spectraldefense}
Paula Harder, Franz-Josef Pfreundt, Margret Keuper, and Janis Keuper.
\newblock Spectraldefense: Detecting adversarial attacks on cnns in the fourier
  domain.
\newblock \emph{arXiv preprint arXiv:2103.03000}, 2021.

\bibitem[He et~al.(2016)He, Zhang, Ren, and Sun]{he2016deep}
Kaiming He, Xiangyu Zhang, Shaoqing Ren, and Jian Sun.
\newblock Deep residual learning for image recognition.
\newblock In \emph{CVPR}, 2016.

\bibitem[Hein and Andriushchenko(2017)]{hein2017formal}
Matthias Hein and Maksym Andriushchenko.
\newblock Formal guarantees on the robustness of a classifier against
  adversarial manipulation.
\newblock In \emph{NeurIPS}, 2017.

\bibitem[Huang et~al.(2017)Huang, Liu, Van Der~Maaten, and
  Weinberger]{huang2017densely}
Gao Huang, Zhuang Liu, Laurens Van Der~Maaten, and Kilian~Q Weinberger.
\newblock Densely connected convolutional networks.
\newblock In \emph{CVPR}, 2017.

\bibitem[Ignatiev et~al.(2019)Ignatiev, Narodytska, and
  Marques-Silva]{ignatiev2019relating}
Alexey Ignatiev, Nina Narodytska, and Joao Marques-Silva.
\newblock On relating explanations and adversarial examples.
\newblock In \emph{NeurIPS}, 2019.

\bibitem[Ilyas et~al.(2018)Ilyas, Engstrom, Athalye, and Lin]{ilyas2018black}
Andrew Ilyas, Logan Engstrom, Anish Athalye, and Jessy Lin.
\newblock Black-box adversarial attacks with limited queries and information.
\newblock In \emph{ICML}, 2018.

\bibitem[Ilyas et~al.(2019)Ilyas, Santurkar, Tsipras, Engstrom, Tran, and
  Madry]{ilyas2019adversarial}
Andrew Ilyas, Shibani Santurkar, Dimitris Tsipras, Logan Engstrom, Brandon
  Tran, and Aleksander Madry.
\newblock Adversarial examples are not bugs, they are features.
\newblock In \emph{NeurIPS}, 2019.

\bibitem[Janizek et~al.(2020)Janizek, Sturmfels, and
  Lee]{janizek2020explaining}
Joseph~D Janizek, Pascal Sturmfels, and Su-In Lee.
\newblock Explaining explanations: Axiomatic feature interactions for deep
  networks.
\newblock \emph{arXiv preprint arXiv:2002.04138}, 2020.

\bibitem[Jere et~al.(2020)Jere, Kumar, and Koushanfar]{jere2020singular}
Malhar Jere, Maghav Kumar, and Farinaz Koushanfar.
\newblock A singular value perspective on model robustness.
\newblock \emph{arXiv preprint arXiv:2012.03516}, 2020.

\bibitem[Jin et~al.(2019)Jin, Wei, Du, Xue, and Ren]{jin2019towards}
Xisen Jin, Zhongyu Wei, Junyi Du, Xiangyang Xue, and Xiang Ren.
\newblock Towards hierarchical importance attribution: Explaining compositional
  semantics for neural sequence models.
\newblock In \emph{ICLR}, 2019.

\bibitem[Kurakin et~al.(2016)Kurakin, Goodfellow, and
  Bengio]{kurakin2016adversarial}
Alexey Kurakin, Ian Goodfellow, and Samy Bengio.
\newblock Adversarial examples in the physical world.
\newblock \emph{arXiv preprint arXiv:1607.02533}, 2016.

\bibitem[Liu et~al.(2017)Liu, Chen, Liu, and Song]{liu2016delving}
Yanpei Liu, Xinyun Chen, Chang Liu, and Dawn Song.
\newblock Delving into transferable adversarial examples and black-box attacks.
\newblock In \emph{ICLR}, 2017.

\bibitem[Lundberg et~al.(2018)Lundberg, Erion, and Lee]{lundberg2018consistent}
Scott~M Lundberg, Gabriel~G Erion, and Su-In Lee.
\newblock Consistent individualized feature attribution for tree ensembles.
\newblock \emph{arXiv preprint arXiv:1802.03888}, 2018.

\bibitem[Ma et~al.(2018)Ma, Li, Wang, Erfani, Wijewickrema, Schoenebeck, Song,
  Houle, and Bailey]{ma2018characterizing}
Xingjun Ma, Bo~Li, Yisen Wang, Sarah~M Erfani, Sudanthi Wijewickrema, Grant
  Schoenebeck, Dawn Song, Michael~E Houle, and James Bailey.
\newblock Characterizing adversarial subspaces using local intrinsic
  dimensionality.
\newblock In \emph{ICLR}, 2018.

\bibitem[Madry et~al.(2018)Madry, Makelov, Schmidt, Tsipras, and
  Vladu]{madry2018towards}
Aleksander Madry, Aleksandar Makelov, Ludwig Schmidt, Dimitris Tsipras, and
  Adrian Vladu.
\newblock Towards deep learning models resistant to adversarial attacks.
\newblock In \emph{ICLR}, 2018.

\bibitem[Meng and Chen(2017)]{meng2017magnet}
Dongyu Meng and Hao Chen.
\newblock Magnet: a two-pronged defense against adversarial examples.
\newblock In \emph{SIGSAC}, 2017.

\bibitem[Murdoch et~al.(2018)Murdoch, Liu, and Yu]{murdoch2018beyond}
W~James Murdoch, Peter~J Liu, and Bin Yu.
\newblock Beyond word importance: Contextual decomposition to extract
  interactions from lstms.
\newblock In \emph{ICLR}, 2018.

\bibitem[Nayebi and Ganguli(2017)]{nayebi2017biologically}
Aran Nayebi and Surya Ganguli.
\newblock Biologically inspired protection of deep networks from adversarial
  attacks.
\newblock \emph{arXiv preprint arXiv:1703.09202}, 2017.

\bibitem[Pal and Vidal(2020)]{pal2020game}
Ambar Pal and Rene Vidal.
\newblock A game theoretic analysis of additive adversarial attacks and
  defenses.
\newblock In \emph{NeurIPS}, 2020.

\bibitem[Papernot et~al.(2016)Papernot, McDaniel, Jha, Fredrikson, Celik, and
  Swami]{papernot2016limitations}
Nicolas Papernot, Patrick McDaniel, Somesh Jha, Matt Fredrikson, Z~Berkay
  Celik, and Ananthram Swami.
\newblock The limitations of deep learning in adversarial settings.
\newblock In \emph{EuroS\&P}, 2016.

\bibitem[Papernot et~al.(2017)Papernot, McDaniel, Goodfellow, Jha, Celik, and
  Swami]{papernot2017practical}
Nicolas Papernot, Patrick McDaniel, Ian Goodfellow, Somesh Jha, Z~Berkay Celik,
  and Ananthram Swami.
\newblock Practical black-box attacks against machine learning.
\newblock In \emph{ASIA-CCS}, 2017.

\bibitem[Qi et~al.(2017)Qi, Su, Mo, and Guibas]{qi2017pointnet}
Charles~R Qi, Hao Su, Kaichun Mo, and Leonidas~J Guibas.
\newblock Pointnet: Deep learning on point sets for 3d classification and
  segmentation.
\newblock In \emph{CVPR}, 2017.

\bibitem[Ren et~al.(2021)Ren, Zhou, Chen, and Zhang]{ren2021learning}
Jie Ren, Zhanpeng Zhou, Qirui Chen, and Quanshi Zhang.
\newblock Learning baseline values for shapley values.
\newblock \emph{arXiv preprint arXiv:2105.10719}, 2021.

\bibitem[Russakovsky et~al.(2015)Russakovsky, Deng, Su, Krause, Satheesh, Ma,
  Huang, Karpathy, Khosla, Bernstein, Berg, and Fei-Fei]{imagenet2015}
Olga Russakovsky, Jia Deng, Hao Su, Jonathan Krause, Sanjeev Satheesh, Sean Ma,
  Zhiheng Huang, Andrej Karpathy, Aditya Khosla, Michael Bernstein,
  Alexander~C. Berg, and Li~Fei-Fei.
\newblock Imagenet large scale visual recognition challenge.
\newblock \emph{In {International Journal of Computer Vision}}, 115\penalty0
  (3):\penalty0 211--252, 2015.

\bibitem[Salman et~al.(2020)Salman, Ilyas, Engstrom, Kapoor, and
  Madry]{salman2020adversarially}
Hadi Salman, Andrew Ilyas, Logan Engstrom, Ashish Kapoor, and Aleksander Madry.
\newblock Do adversarially robust imagenet models transfer better?
\newblock \emph{arXiv preprint arXiv:2007.08489}, 2020.

\bibitem[Shapley(1953)]{shapley1953value}
Lloyd~S Shapley.
\newblock A value for n-person games.
\newblock \emph{Contributions to the Theory of Games}, 2\penalty0
  (28):\penalty0 307--317, 1953.

\bibitem[Simonyan and Zisserman(2015)]{simonyan2015very}
Karen Simonyan and Andrew Zisserman.
\newblock Very deep convolutional networks for large-scale image recognition.
\newblock In \emph{ICLR}, 2015.

\bibitem[Singh et~al.(2018)Singh, Murdoch, and Yu]{singh2018hierarchical}
Chandan Singh, W~James Murdoch, and Bin Yu.
\newblock Hierarchical interpretations for neural network predictions.
\newblock In \emph{ICLR}, 2018.

\bibitem[Song et~al.(2018)Song, Shu, Kushman, and Ermon]{song2018constructing}
Yang Song, Rui Shu, Nate Kushman, and Stefano Ermon.
\newblock Constructing unrestricted adversarial examples with generative
  models.
\newblock \emph{NeurIPS}, 2018.

\bibitem[Sorokina et~al.(2008)Sorokina, Caruana, Riedewald, and
  Fink]{sorokina2008detecting}
Daria Sorokina, Rich Caruana, Mirek Riedewald, and Daniel Fink.
\newblock Detecting statistical interactions with additive groves of trees.
\newblock In \emph{ICML}, 2008.

\bibitem[Srivastava et~al.(2014)Srivastava, Hinton, Krizhevsky, Sutskever, and
  Salakhutdinov]{srivastava2014dropout}
Nitish Srivastava, Geoffrey Hinton, Alex Krizhevsky, Ilya Sutskever, and Ruslan
  Salakhutdinov.
\newblock Dropout: a simple way to prevent neural networks from overfitting.
\newblock \emph{The journal of Machine Learning Research}, 15\penalty0
  (1):\penalty0 1929--1958, 2014.

\bibitem[Sundararajan et~al.(2017)Sundararajan, Taly, and
  Yan]{sundararajan2017axiomatic}
Mukund Sundararajan, Ankur Taly, and Qiqi Yan.
\newblock Axiomatic attribution for deep networks.
\newblock In \emph{ICML}, 2017.

\bibitem[Sundararajan et~al.(2020)Sundararajan, Dhamdhere, and
  Agarwal]{sundararajan2020shapley}
Mukund Sundararajan, Kedar Dhamdhere, and Ashish Agarwal.
\newblock The shapley taylor interaction index.
\newblock In \emph{ICML}, 2020.

\bibitem[Szegedy et~al.(2013)Szegedy, Zaremba, Sutskever, Bruna, Erhan,
  Goodfellow, and Fergus]{szegedy2013intriguing}
Christian Szegedy, Wojciech Zaremba, Ilya Sutskever, Joan Bruna, Dumitru Erhan,
  Ian Goodfellow, and Rob Fergus.
\newblock Intriguing properties of neural networks.
\newblock \emph{arXiv preprint arXiv:1312.6199}, 2013.

\bibitem[Tian et~al.(2021)Tian, Kuang, Jiang, Wu, and Wang]{tian2021analysis}
Qi~Tian, Kun Kuang, Kelu Jiang, Fei Wu, and Yisen Wang.
\newblock Analysis and applications of class-wise robustness in adversarial
  training.
\newblock In \emph{KDD}, 2021.

\bibitem[Tsang et~al.(2018)Tsang, Cheng, and Liu]{tsang2018detecting}
Michael Tsang, Dehua Cheng, and Yan Liu.
\newblock Detecting statistical interactions from neural network weights.
\newblock In \emph{ICLR}, 2018.

\bibitem[Tsipras et~al.(2018)Tsipras, Santurkar, Engstrom, Turner, and
  Madry]{tsipras2018robustness}
Dimitris Tsipras, Shibani Santurkar, Logan Engstrom, Alexander Turner, and
  Aleksander Madry.
\newblock Robustness may be at odds with accuracy.
\newblock In \emph{ICLR}, 2018.

\bibitem[Wang et~al.(2020{\natexlab{a}})Wang, Wu, Huang, and
  Xing]{wang2020high}
Haohan Wang, Xindi Wu, Zeyi Huang, and Eric~P Xing.
\newblock High-frequency component helps explain the generalization of
  convolutional neural networks.
\newblock In \emph{CVPR}, 2020{\natexlab{a}}.

\bibitem[Wang et~al.(2020{\natexlab{b}})Wang, Ren, Lin, Zhu, Wang, and
  Zhang]{wang2020unified}
Xin Wang, Jie Ren, Shuyun Lin, Xiangming Zhu, Yisen Wang, and Quanshi Zhang.
\newblock A unified approach to interpreting and boosting adversarial
  transferability.
\newblock \emph{arXiv preprint arXiv:2010.04055}, 2020{\natexlab{b}}.

\bibitem[Wang et~al.(2021)Wang, Lin, Zhang, Zhu, and
  Zhang]{wang2021interpreting}
Xin Wang, Shuyun Lin, Hao Zhang, Yufei Zhu, and Quanshi Zhang.
\newblock Interpreting attributions and interactions of adversarial attacks.
\newblock In \emph{ICCV}, 2021.

\bibitem[Wang et~al.(2019)Wang, Ma, Bailey, Yi, Zhou, and Gu]{wang2019dynamic}
Yisen Wang, Xingjun Ma, James Bailey, Jinfeng Yi, Bowen Zhou, and Quanquan Gu.
\newblock On the convergence and robustness of adversarial training.
\newblock In \emph{ICML}, 2019.

\bibitem[Wang et~al.(2020{\natexlab{c}})Wang, Zou, Yi, Bailey, Ma, and
  Gu]{wang2020improving}
Yisen Wang, Difan Zou, Jinfeng Yi, James Bailey, Xingjun Ma, and Quanquan Gu.
\newblock Improving adversarial robustness requires revisiting misclassified
  examples.
\newblock In \emph{ICLR}, 2020{\natexlab{c}}.

\bibitem[Wang et~al.(2018)Wang, Jha, and Chaudhuri]{wang2018analyzing}
Yizhen Wang, Somesh Jha, and Kamalika Chaudhuri.
\newblock Analyzing the robustness of nearest neighbors to adversarial
  examples.
\newblock In \emph{ICML}, 2018.

\bibitem[Weber(1988)]{weber1988probabilistic}
Robert~J Weber.
\newblock Probabilistic values for games.
\newblock \emph{The Shapley Value. Essays in Honor of Lloyd S. Shapley}, pages
  101--119, 1988.

\bibitem[Weng et~al.(2018)Weng, Zhang, Chen, Yi, Su, Gao, Hsieh, and
  Daniel]{weng2018evaluating}
Tsui-Wei Weng, Huan Zhang, Pin-Yu Chen, Jinfeng Yi, Dong Su, Yupeng Gao,
  Cho-Jui Hsieh, and Luca Daniel.
\newblock Evaluating the robustness of neural networks: An extreme value theory
  approach.
\newblock In \emph{ICLR}, 2018.

\bibitem[Wu et~al.(2020{\natexlab{a}})Wu, Wang, Xia, Bailey, and
  Ma]{wu2020skip}
Dongxian Wu, Yisen Wang, Shu-Tao Xia, James Bailey, and Xingjun Ma.
\newblock Skip connections matter: On the transferability of adversarial
  examples generated with resnets.
\newblock In \emph{ICLR}, 2020{\natexlab{a}}.

\bibitem[Wu et~al.(2020{\natexlab{b}})Wu, Xia, and Wang]{wu2020adversarial}
Dongxian Wu, Shu-Tao Xia, and Yisen Wang.
\newblock Adversarial weight perturbation helps robust generalization.
\newblock In \emph{NeurIPS}, 2020{\natexlab{b}}.

\bibitem[Wu et~al.(2014)Wu, Song, Khosla, Tang, and Xiao]{wu20143d}
Zhirong Wu, Shuran Song, Aditya Khosla, Xiaoou Tang, and Jianxiong Xiao.
\newblock 3d shapenets for 2.5 d object recognition and next-best-view
  prediction.
\newblock \emph{arXiv preprint arXiv:1406.5670}, 2014.

\bibitem[Xie et~al.(2019)Xie, Wu, Maaten, Yuille, and He]{xie2019feature}
Cihang Xie, Yuxin Wu, Laurens van~der Maaten, Alan~L Yuille, and Kaiming He.
\newblock Feature denoising for improving adversarial robustness.
\newblock In \emph{CVPR}, 2019.

\bibitem[Xu et~al.(2018)Xu, Liu, Zhao, Chen, Zhang, Fan, Erdogmus, Wang, and
  Lin]{xu2018structured}
Kaidi Xu, Sijia Liu, Pu~Zhao, Pin-Yu Chen, Huan Zhang, Quanfu Fan, Deniz
  Erdogmus, Yanzhi Wang, and Xue Lin.
\newblock Structured adversarial attack: Towards general implementation and
  better interpretability.
\newblock In \emph{ICLR}, 2018.

\bibitem[Xu et~al.(2019)Xu, Liu, Zhang, Sun, Zhao, Fan, Gan, and
  Lin]{xu2019interpreting}
Kaidi Xu, Sijia Liu, Gaoyuan Zhang, Mengshu Sun, Pu~Zhao, Quanfu Fan, Chuang
  Gan, and Xue Lin.
\newblock Interpreting adversarial examples by activation promotion and
  suppression.
\newblock \emph{arXiv preprint arXiv:1904.02057}, 2019.

\bibitem[Yang et~al.(2019)Yang, Chen, Hsieh, Wang, and
  Jordan]{DBLP:journals/corr/abs-1906-03499}
Puyudi Yang, Jianbo Chen, Cho{-}Jui Hsieh, Jane{-}Ling Wang, and Michael~I.
  Jordan.
\newblock {ML-LOO:} detecting adversarial examples with feature attribution.
\newblock \emph{arXiv preprint arXiv:1906.03499}, 2019.

\bibitem[Yin et~al.(2019)Yin, Lopes, Shlens, Cubuk, and Gilmer]{yin2019fourier}
Dong Yin, Raphael~Gontijo Lopes, Jonathon Shlens, Ekin~D Cubuk, and Justin
  Gilmer.
\newblock A fourier perspective on model robustness in computer vision.
\newblock \emph{arXiv preprint arXiv:1906.08988}, 2019.

\bibitem[Zagoruyko and Komodakis(2016)]{zagoruyko2016wide}
Sergey Zagoruyko and Nikos Komodakis.
\newblock Wide residual networks.
\newblock \emph{arXiv preprint arXiv:1605.07146}, 2016.

\bibitem[Zhang et~al.(2021{\natexlab{a}})Zhang, Zhou, Zhang, Bao, Huo, Chen,
  Cheng, Wu, and Zhang]{zhang2021building}
Die Zhang, Huilin Zhou, Hao Zhang, Xiaoyi Bao, Da~Huo, Ruizhao Chen, Xu~Cheng,
  Mengyue Wu, and Quanshi Zhang.
\newblock Building interpretable interaction trees for deep nlp models.
\newblock In \emph{AAAI}, 2021{\natexlab{a}}.

\bibitem[Zhang et~al.(2020)Zhang, Cheng, Chen, and Zhang]{zhang2020game}
Hao Zhang, Xu~Cheng, Yiting Chen, and Quanshi Zhang.
\newblock Game-theoretic interactions of different orders.
\newblock \emph{arXiv preprint arXiv:2010.14978}, 2020.

\bibitem[Zhang et~al.(2021{\natexlab{b}})Zhang, Li, Ma, Li, Xie, and
  Zhang]{zhang2020interpreting-dropout}
Hao Zhang, Sen Li, Yinchao Ma, Mingjie Li, Yichen Xie, and Quanshi Zhang.
\newblock Interpreting and boosting dropout from a game-theoretic view.
\newblock In \emph{ICLR}, 2021{\natexlab{b}}.

\bibitem[Zhang et~al.(2021{\natexlab{c}})Zhang, Xie, Zheng, Zhang, and
  Zhang]{zhang2021interpreting}
Hao Zhang, Yichen Xie, Longjie Zheng, Die Zhang, and Quanshi Zhang.
\newblock Interpreting multivariate interactions in dnns.
\newblock In \emph{AAAI}, 2021{\natexlab{c}}.

\bibitem[Zhang et~al.(2019)Zhang, Yu, Jiao, Xing, El~Ghaoui, and
  Jordan]{zhang2019theoretically}
Hongyang Zhang, Yaodong Yu, Jiantao Jiao, Eric Xing, Laurent El~Ghaoui, and
  Michael Jordan.
\newblock Theoretically principled trade-off between robustness and accuracy.
\newblock In \emph{ICML}, 2019.

\bibitem[Zhang and Zhu(2019)]{zhang2019interpreting}
Tianyuan Zhang and Zhanxing Zhu.
\newblock Interpreting adversarially trained convolutional neural networks.
\newblock In \emph{ICML}, 2019.

\end{thebibliography}
	}
	
	
\newpage
\appendix

\section{Preliminaries: Shapley values}
	
In this section, in order to help readers understand the metric in the paper, we first revisit the definition  of the Shapley value~\cite{shapley1953value}, which is widely considered as an unbiased estimation of the numerical importance \emph{w.r.t.} each input variable.
In game theory, the complex system is usually represented as a game, where each input variable is taken as a player, and the output of this system is regarded as the total reward of all players. Given a game with multiple players (input variables) $N=\{1,2,\cdots,n\}$, some players cooperate to pursue a high reward.
Thus, the task is to divide the total reward, and fairly assign the divided elementary reward to each individual player. In this way, the elementary reward can be considered as the numerical importance of the corresponding variable to the complex system. Let $2^N \overset{\textrm{def}}{=} \{S|S\subseteq N\}$ indicate all potential subsets of $N$. The game $v:2^{N} \rightarrow \mathbb{R}$ is a function, which estimates the overall reward $v(S)$ earned by each specific subset of players $S\subseteq N$. In this way, the Shapley value, denoted by $\phi(i)$, represents the numerical importance of the player $i$ to the game $v$.
\begin{equation}
\label{eqn:shapleyvalue}
\phi(i)=\!\!\!\sum_{S\subseteq N\setminus\{i\}}\!\!\!\frac{(n-|S|-1)!|S|!}{n!}\Big[v(S\cup\{i\})-v(S)\Big].
\end{equation}
\citet{weber1988probabilistic} has proven that the Shapley value is a unique method to fairly allocate overall reward to each player that satisfies following properties.

\textbf{(1) Linearity property}: If two independent games can be merged into one game $u(S)=v(S)+w(S)$, then the Shapley value of the new game also can be merged, \emph{i.e.} $\forall i \in N$, $\phi_{u}(i)=\phi_{v}(i)+\phi_{w}(i)$; $\forall c \in \mathbb{R}$, $\phi_{c \cdot u} (i)= c\cdot \phi_{u}(i)$.

\textbf{(2) Nullity property}: The dummy player $i$ is defined as a player satisfying $\forall S\subseteq N\setminus\{i\}$, $v(S\cup\{i\})=v(S)+v(\{i\})$, which indicates that the player $i$ has no interactions with other players in $N$, $\phi(i)=v(\{i\})- v(\emptyset)$.

\textbf{(3) Symmetry property}: If $\forall S\subseteq N\setminus\{i,j\}$, $v(S\cup\{i\})=v(S\cup\{j\})$, then $\phi(i)=\phi(j)$.

\textbf{(4) Efficiency property}: The overall reward can be assigned to all players, $\sum_{i\in N}\phi(i)=v(N) - v(\emptyset)$.

\textbf{Using Shapley values to explain DNNs.}
Given a trained DNN and the input with $n$ variables $N=\{1,\cdots,n\}$, we can take the input variables as players, and consider the network output as the reward.
The Shapley value $\phi(i)$ of each input variable $i\in N$ is regarded as the importance of the variable $i$ to the network output.
Each subset of variables $S\subseteq N$ represents a specific context.
$v(S)$ represents the network output, when we keep variables in $S$ unchanged and mask variables in $N\setminus S$ by following settings in \cite{ancona2019explaining}.
In particular, $v(N)$ refers to the network output \emph{w.r.t.} the entire input $N$, and $v(\emptyset)$ denotes the output when we mask all variables.
Note that for the DNN trained for multi-category classification, $v(S)$ can be taken as an arbitrary dimension of the network output, so as to measure the variable importance to the corresponding category.

\section{Multi-order interactions and multi-order Shapley values}

In this section, we first revisit the definition of the Shapley interaction index~\cite{grabisch1999axiomatic} and multi-order interactions~\cite{zhang2020interpreting-dropout}, in order to improve the readability.
Then, we provide proofs of extended properties of multi-order interactions, proofs of the relationship between multi-order interactions and multi-order Shapley values, and the proof of Proposition 1 in Section 4.1 of the paper.

\subsection{Multi-order interactions}

\paragraph{Shapley interaction index.}
Input variables of a DNN do not contribute to the network output independently.
Instead, there are interactions/collaborations between different variables.
To this end, the Shapley interaction index~\cite{grabisch1999axiomatic} measures the influence of a variable on the Shapley value (importance) of another variable.
\emph{I.e.} for two variables $(i,j)$,  it examines whether the absence/presence of $j$ can change the importance of $i$.
Thus, the Shapley interaction index is defined as the change in the Shapley value (importance) of variable $i$ when the variable $j$ is always present \emph{w.r.t.} the case when $j$ is always absent, as follows.
\begin{equation}
I(i,j)=\tilde{\phi}(i)_{\text{$j$ always present}} - \tilde{\phi}(i)_{\text{$j$ always absent}}
\end{equation}
where $\tilde{\phi}(i)_{\text{$j$ always present}}$ denotes the Shapley value of the variable $i$ computed under the specific condition that $j$ is always present.
$\tilde{\phi}(i)_{\text{$j$ always absent}}$ is computed under the specific condition that $j$ is always absent.
Note that $I(i,j) = I(j,i)$.
If $I(i,j)>0$, it indicates that the presence of the variable $j$ will increase the importance of the variable $i$. Thus, we consider variables $(i,j)$ have a positive interaction.
If $I(i,j) < 0$, it indicates a negative interaction.

\paragraph{Multi-order interactions.}
\citet{zhang2020interpreting-dropout} decomposed the Shapley interaction index into interactions of different orders, as follows.
\begin{equation}
\label{eqn:multi-order_interaction}
I^{(m)}_{ij} = \mathbb{E}_{S\subseteq N\backslash \{i,j\},|S|=m}[\Delta v(i,j,S)],
\end{equation}
where $\Delta v(i,j,S) \overset{\textrm{def}}{=} v(S \cup \{i,j\})-v(S \cup \{i\}) - v(S\cup \{j\}) + v(S)$.
$I_{ij}^{(m)}$ denotes the interaction of the $m$-th order, which measures the average interaction between variables $(i,j)$ under all contexts consisting of $m$ variables (\emph{e.g.} a visual context consisting of $m$ pixels).
For a low order $m$, $I^{(m)}_{ij}$ reflects the interaction between $i$ and $j$ \emph{w.r.t.} simple contextual collaborations with a few variables.
For a high order $m$, $I^{(m)}_{ij}$ corresponds to the interaction \emph{w.r.t.} complex contextual collaborations with massive variables.
%
%

\textbf{Proofs of new properties of multi-order interactions claimed in the paper.}
In Section 3 of the paper, we claim that $I^{(m)}_{ij}$ satisfies \emph{linearity, nullity, commutativity, symmetry,} and \emph{efficiency} properties.
Proofs are given as follows.

\textbf{(1) Linearity property}:
If we merge outputs of two DNNs, $u(S)=w(S)+v(S)$, then, $\forall i,j\in N$, the interaction $I^{(m)}_{ij,u}$ \emph{w.r.t.} the new output $u$ can be decomposed into $I^{(m)}_{ij,u} = I^{(m)}_{ij,w} + I^{(m)}_{ij,v}$.

$\bullet\;$\emph{Proof}:
\begin{align*}
I^{(m)}_{ij,u} &=\mathbb{E}_{S\subseteq N\backslash \{i,j\}, |S| = m} [ \Delta u(S,i,j)]\\
&= \mathbb{E}_{S\subseteq N\backslash \{i,j\}, |S| = m} [\Delta v(i,j,S) +\Delta w(S,i,j) ]\\
&= I^{(m)}_{ij,v} +  I^{(m)}_{ij,w}
\end{align*}$\hfill\square$

\textbf{(2) Nullity property}:
The dummy variable $i\in N$ satisfies $\forall S\subseteq N\setminus\{i\}$, $v(S\cup \{i\}) = v(S)+v(\{i\})$. It means that the variable $i$ has no interactions with other variables, \emph{i.e.} $\forall j \in N$, $I^{(m)}_{ij}=0$.

$\bullet\;$\emph{Proof}:
\begin{align*}
I^{(m)}_{ij} &= \mathbb{E}_{S\subseteq N\backslash \{i,j\}, |S| = m} [ v(S\cup \{i,j\}) - v(S\cup \{i\}) - v(S\cup \{j\}) + v(S)]\\
& = \mathbb{E}_{S\subseteq N\backslash \{i,j\}, |S| = m} \big[ v(S\cup \{j\} \cup \{i\})- v(S\cup \{j\} ) - \big(v(S\cup \{i\}) - v(S)\big)\big]\\
& =\mathbb{E}_{S\subseteq N\backslash \{i,j\}, |S| = m} [ v(\{i\}) - v(\{i\})]
= 0
\end{align*}$\hfill\square$

\textbf{(3) Commutativity property}:
$\forall i,j \in N$, $ I^{(m)}_{ij} = I^{(m)}_{ji}$.

$\bullet\;$\emph{Proof}:
\begin{align*}
I^{(m)}_{ij} &= \mathbb{E}_{S\subseteq N\backslash \{i,j\}, |S| = m} [v(S\cup\{i,j\})-v(S\cup\{i\})-v(S\cup\{j\})+v(S)]\\
&= \mathbb{E}_{S\subseteq N\backslash \{i,j\}, |S| = m} [v(S\cup\{i,j\})-v(S\cup\{j\})-v(S\cup\{i\})+v(S)]\\
&=I^{(m)}_{ji}
\end{align*}$\hfill\square$

\textbf{(4) Symmetry property}:
If input variables $i,j \in N$ have same cooperations with other variables $\forall S\subseteq N \backslash \{i,j\}$, $v(S\cup \{i\}) = v(S\cup \{j\})$, then they have same interactions, $\forall k \in N\backslash\{i,j\}$, $ I^{(m)}_{ik} = I^{(m)}_{jk}$.


$\bullet\;$\emph{Proof}:
\begin{align*}
I^{(m)}_{ik} & = \mathbb{E}_{\substack{S\subseteq N\backslash \{i,j\}, \\|S| = m}} [\Delta v(i,k,S)]\\
& = \frac{m!(n-2-m)!}{(n-2)!}\sum_{\substack{S\subseteq N\backslash \{i,k\},\\ |S| = m}} [ \Delta v(i,k,S)]\\
& = \frac{m!(n-2-m)!}{(n-2)!}\left(\sum_{\substack{S\subseteq N\backslash \{i,j,k\},\\ |S| = m-1}} [ \Delta v(i,k,S\cup \{j\})]+\sum_{\substack{S\subseteq N\backslash \{i,j,k\},\\ |S| = m}} [\Delta  v(i,k,S)]\right)\\
& = \frac{m!(n-2-m)!}{(n-2)!}\left(\sum_{\substack{S\subseteq N\backslash \{i,j,k\},\\ |S| = m-1}} [ \Delta v(j,k,S\cup \{i\})]+\sum_{\substack{S\subseteq N\backslash \{i,j,k\},\\ |S| = m}} [\Delta  v(j,k,S)]\right)\\
& = \mathbb{E}_{\substack{S\subseteq N\backslash \{i,j\}, \\|S| = m}} [\Delta v(j,k,S)]\\
& = I^{(m)}_{jk}
\end{align*}$\hfill\square$

\textbf{(5) Efficiency property}:
The output of the DNN can be decomposed into interactions of different orders, $v(N)=v(\emptyset)+\sum_{i\in N}\phi^{(0)}(i) +\sum_{i\in N}\sum_{j\in N\backslash \{i\}} [\sum_{m=0}^{n-2}\frac{n-1-m}{n(n-1)}I^{(m)}_{ij}]$, where $\phi^{(0)}(i)\overset{\textrm{def}}{=}v(i)-v(\emptyset)$.

$\bullet\;$\emph{Proof}:
\begin{align*}
v(N)&=v(\emptyset)+ \frac{1}{n}\sum\nolimits_{i\in N}\sum\nolimits_{m=0}^{n-1}\phi^{(m)}(i)\\
&=v(\emptyset) +\frac{1}{n}\sum\nolimits_{i\in N}\phi^{(0)}(i)+ \frac{1}{n}\sum\nolimits_{i\in N}\sum\nolimits_{m=1}^{n-1} \Big[ \mathbb{E}_{j \in N\backslash\{i\}}\big[\sum\nolimits_{k=0}^{k=m-1}I^{(k)}_{ij}\big]+ \phi^{(0)}(i)\Big]\\
&= v(\emptyset)+ \sum\nolimits_{i\in N}\phi^{(0)}(i)+\frac{1}{n(n-1)}\sum\nolimits_{i\in N}\sum\nolimits_{j\in N\backslash \{i\}}\big[\sum\nolimits_{m=1}^{n-1}\sum\nolimits_{k=0}^{k=m-1}I^{(k)}_{ij}\big]\\
&= v(\emptyset) + \sum\nolimits_{i\in N}\phi^{(0)}(i)+\sum\nolimits_{i\in N}\sum\nolimits_{j\in N\backslash \{i\}} [\sum\nolimits_{m=0}^{n-2}\frac{n-1-m}{n(n-1)}I^{(m)}_{ij}]
\end{align*}$\hfill\square$

\subsection{Multi-order Shapley values}
In Section 4.3  of the paper, we define multi-order Shapley values.
In the supplementary material, this section provides more details about multi-order Shapley values  to help readers understand, and also provides proofs of extended properties of multi-order Shapley values.

We decompose the Shapley value $\phi(i)$ into different orders, as follows.
\begin{flalign}
\phi(i)&= \frac{1}{n}\sum\nolimits_{m=0}^{n-1}\phi^{(m)}(i), \label{eqn:multi_order_shapley_sum}\\
\phi^{(m)}(i)&=\mathbb{E}_{S\subseteq N\setminus\{i\},|S|=m} [ v(S\cup {i})-v(S)], \label{eqn:multi_order_shapley}
\end{flalign}
where $\phi^{(m)}(i)$ denotes the Shapley value of the $m$-th order. It measures the importance of the input variable $i$ to the network output with contexts consisting of $m\in \{0,\dots,n-1\}$ variables.

For a low order $m$, $\phi^{(m)}(i)$ denotes the importance of the variable $i$, when $i$ cooperates with a few contextual variables for inference. For a high order $m$, $\phi^{(m)}(i)$ describes the importance of the variable $i$, which cooperates with massive contextual variables. In particular, $\phi^{(0)}(i)= v({i})-v(\emptyset)$ represents the numerical importance of $i$ without taking into account any contexts.

In addition, we have proven that $\phi^{(m)}(i)$ satisfies following properties.

\textbf{(1) Linearity property}:
If we merge the outputs of two DNNs $u(S)=w(S)+v(S)$, then the Shapley values of input variables also can be added, \emph{i.e.}   $\forall i\in N$, $\phi_u^{(m)}(i) = \phi_w^{(m)}(i) + \phi_v^{(m)}(i)$.

$\bullet\;$\emph{Proof}:
\begin{small}
\begin{align*}
	\phi_u^{(m)}(i) & = \mathbb{E}_{S\subseteq N\backslash \{i\}, |S| = m} [ u(S\cup \{i\}) - u(S)]\\
	& =\mathbb{E}_{S\subseteq N\backslash \{i\},  |S| = m} [ w(S\cup \{i\}) + v(S\cup \{i\}) - w(S) - v(S)]\\
	& =\mathbb{E}_{S\subseteq N\backslash \{i\}, |S| = m} [ w(S\cup \{i\})- w(S)] + \mathbb{E}_{S\subseteq N\backslash \{i\},|S| = m} [ v(S\cup \{i\})- v(S)] \\
	& = \phi_w^{(m)}(i) + \phi_v^{(m)}(i)
\end{align*}
\end{small}
$\hfill\square$

\textbf{(2) Nullity property}:
An input variable $i\in N$ is considered as a dummy player if $\forall S\subseteq N\setminus\{i\}$, $v(S\cup \{i\}) = v(S)+v(\{i\})$. Thus, the variable $i$ has no interactions with other variables, \emph{i.e.} $\phi^{(m)}(i)=v(\{i\})$.

$\bullet\;$\emph{Proof}:
\begin{small}
\begin{align*}
	\phi^{(m)}(i) & = \mathbb{E}_{S\subseteq N\backslash \{i\}, |S| = m} [ v(S\cup \{i\}) - v(S)]\\
	& = \mathbb{E}_{S\subseteq N\backslash \{i\}, |S| = m} [ v(\{i\})]
	= v(\{i\})
\end{align*}
\end{small}
$\hfill\square$

\textbf{(3) Symmetry property}:
Given two input variables $i,j\in N$, if these two variables have same cooperations with all other variables $\forall S\subseteq N \backslash \{i,j\}$, $v(S\cup \{i\}) = v(S\cup \{j\})$, then $ \phi^{(m)}(i) = \phi^{(m)}(j) $.


$\bullet\;$\emph{Proof}:
\begin{small}
\begin{align*}
	\phi^{(m)}(i) & = \mathbb{E}_{S\subseteq N\backslash \{i\}, |S| = m} [ v(S\cup \{i\}) - v(S)]\\
	& = \frac{m!(n-1-m)!}{(n-1)!} \!\! \sum_{\substack{S\subseteq N\backslash \{i\},\\ |S| = m}}\!\! [ v(S\cup \{i\}) - v(S)]\\
	& = \frac{m!(n-1-m)!}{(n-1)!}\bigg(\sum_{\substack{S\subseteq N\backslash \{i,j\},\\ |S| = m-1}}\!\! [ v(S\cup \{i,j\}) - v(S,j)]+\!\!\sum_{
		\substack{S\subseteq N\backslash \{i,j\}, \\|S| = m}}\!\! [ v(S\cup \{i\}) - v(S)]\bigg)\\
	& = \frac{m!(n-1-m)!}{(n-1)!}\bigg(\sum_{\substack{S\subseteq N\backslash \{i,j\}, \\|S| = m-1}} \!\![ v(S\cup \{i,j\}) - v(S,i)]+\!\!\sum_{\substack{S\subseteq N\backslash \{i,j\},\\ |S| = m}}\!\! [ v(S\cup \{j\}) - v(S)]\bigg)\\
	& = \mathbb{E}_{S\subseteq N\backslash \{j\}, |S| = m} [ v(S\cup \{j\}) - v(S)]\\
	& = \phi^{(m)}(j)
\end{align*}
\end{small}
$\hfill\square$

\textbf{(4) Efficiency property}: The overall reward can be assigned to all players, $\frac{1}{n}\sum\nolimits_{i\in N}\sum\nolimits_{m=0}^{n-1}\phi^{(m)}(i) = v(N) - v(\emptyset)$.

$\bullet\;$\emph{Proof}:
\begin{small}
\begin{align*}
	v(N) - v(\emptyset) & =\sum\nolimits_{i\in N}\phi(i)
	=\frac{1}{n}\sum\nolimits_{i\in N}\sum\nolimits_{m=0}^{n-1}\phi^{(m)}(i)
\end{align*}
\end{small}
$\hfill\square$

\subsection{Relationship between multi-order Shapley values and multi-order interactions}
This section provides proofs of the relationship between multi-order Shapley values and multi-order interactions.
We have proven that multi-order Shapley values and multi-order interactions satisfy the following \textit{marginal attribution} and \textit{accumulation} properties.

\textbf{(1) Marginal attribution property}:
The marginal attribution of the $(m+1)$-th order Shapley values beyond the $m$-th order is equal to the average interaction of the $m$-th order between $i$ and all other variables. $\forall i,j\in N, i\neq j$, $\phi^{(m+1)}(i)-\phi^{(m)}(i)=\mathbb{E}_{j\in N\backslash \{i\}} [I^{(m)}_{ij}]$.

$\bullet\;$\emph{Proof}:
\begin{small}
\begin{align*}
	\phi^{(m+1)}(i)-\phi^{(m)}(i) =& \mathbb{E}_{S' \subseteq N\backslash\{i\}\atop|S'| = m+1 }\big[v(S'\cup \{i\})-v(S')\big] -\mathbb{E}_{S \subseteq N\backslash\{i\}\atop|S| = m}\big[v(S\cup \{i\})-v(S)\big]\\
	=& \mathbb{E}_{S \subseteq N\backslash\{i\}\atop|S| = m}\Big[\mathbb{E}_{j\in N\backslash(S\cup \{i\})}\big[v(S\cup\{j\}\cup\{i\})\\
	&- v(S\cup\{j\})\big]\Big]-\mathbb{E}_{S \subseteq N\backslash\{i\}\atop|S| = m}\big[v(S\cup \{i\})-v(S)\big]\\
	=&\mathbb{E}_{S \subseteq N\backslash\{i\}\atop|S| = m}\Big[\mathbb{E}_{j\in N\backslash(S\cup \{i\})}\big[v(S\cup\{j\}\cup\{i\})- v(S\cup\{j\})-v(S\cup \{i\})+v(S)\big]\Big]\\
	=& \mathbb{E}_{S \subseteq N\backslash\{i\}\atop|S| = m}\mathbb{E}_{j\in N\backslash(S\cup \{i\})} \big[\Delta v(i,j,S)\big]\\
	=& \mathbb{E}_{j\in N\backslash \{i\}}\mathbb{E}_{S \subseteq N\backslash\{i,j\}\atop|S| = m} \big[\Delta v(i,j,S)\big]\\
	=& \mathbb{E}_{j\in N\backslash \{i\}} [I^{(m)}_{ij}]
\end{align*}
\end{small}
$\hfill\square$

\textbf{(2) Accumulation property}:
The $m$-th ($m>0$) order Shapley value of the variable $i\in N$ can be decomposed into interactions of lower orders, $\phi^{(m)}(i) = \mathbb{E}_{j\in N\backslash\{i\}} [\sum_{k=0}^{m-1} I^{(k)}_{ij}]+ \phi^{(0)}(i)$.

$\bullet\;$\emph{Proof}:
\begin{small}
\begin{align*}
	\phi^{(m)}(i) &= \phi^{(m)}(i) - \phi^{(m-1)}(i) + \phi^{(m-1)}(i) - \phi^{(m-2)}(i)+ \cdots -\phi^{(0)}(i) +\phi^{(0)}(i)\\
	&= \mathbb{E}_{j \in N\backslash\{i\}}[I^{(m-1)}_{ij}]+\mathbb{E}_{j \in N\backslash\{i\}}[I^{(m-2)}_{ij}]+\cdots +\mathbb{E}_{j \in N\backslash\{i\}}[I^{(0)}_{ij}]+ \phi^{(0)}(i)\\
	&=  \mathbb{E}_{j \in N\backslash\{i\}}\Big[\sum\nolimits_{k=0}^{k=m-1}I^{(k)}_{ij}\Big]+ \phi^{(0)}(i)
\end{align*}
\end{small}
$\hfill\square$

\subsection{Equivalence between the multi-order interaction and the mutual information}
\label{sec:equivalence}

Proposition 1 in Section 4.1 of the paper shows the equivalence between the multi-order interaction and the mutual information.
In the supplementary material, this section provides the proof of this proposition.

When a DNN outputs a probability distribution, we prove that the interaction between input variables can be represented in the form of mutual information.
Without loss of generality, let us take the image classification task for example.
Let $x\in X \subseteq \mathbb{R}^n$ denote an input image of the DNN.
$x_i$ denotes the $i$-th pixel, and $X_i=\{x_i\}$.
$\forall S\subseteq N$,
we define $X_S=\{x_S|x\in X\}$; each $x_S$ represents the image, where pixels in $S$ remain unchanged, and other pixels $j\in N\setminus S$ are masked following settings of \cite{ancona2019explaining}. Let $y\in Y=\{y^1,\cdots,y^C\}$ denote the network prediction.
In this way, given $x_S$ as the input, $p(y|x_S)$ denotes the output probability of the DNN.
Let us set {\small$v(S) = H(Y|X_S)=\sum_{x_S} p(x_S) H(Y|X_S= x_S)$}, which measures the entropy of $y$ given the input $x_S$. Then we prove that
\begin{equation}
\begin{small}
	\begin{aligned}
		I^{(m)}_{ij} =\mathbb{E}_{S\subseteq N\setminus\{i,j\},|S|=m} MI(X_i;X_j;Y|X_S)
	\end{aligned}
	\label{eq:equivalence}
\end{small}
\end{equation}

$\bullet\;$\emph{Proof}:
\begin{small}
\begin{align*}
	I^{(m)}_{ij} &= \mathbb{E}_{S\subseteq N\setminus\{i,j\},|S|=m} \Big[v(S\cup \{i,j\}) - v(S\cup \{i\}) - v(S\cup \{j\}) + v(S)\Big]\\
	&= \mathbb{E}_{S\subseteq N\setminus\{i,j\},|S|=m} \Big[-H(Y|X_S,X_i,X_j)+H(Y|X_S,X_i)+H(Y|X_S,X_j)-H(H|X_S)\Big]\\
	&= \mathbb{E}_{S\subseteq N\setminus\{i,j\},|S|=m} \Big[H(Y|X_S,X_j)-H(Y|X_S,X_j,X_i) + H(Y|X_S,X_i)-H(Y|X_S)\Big]\\
	&= \mathbb{E}_{S\subseteq N\setminus\{i,j\},|S|=m} \Big[MI(X_i;Y|X_S,X_j)-MI(X_i;Y|X_S)\Big]\\
	&= \mathbb{E}_{S\subseteq N\setminus\{i,j\},|S|=m} \Big[MI(X_i;X_j;Y|X_S)\Big]
\end{align*}
\end{small}
$\hfill\square$

The conditional mutual information $MI(X_i;X_j;Y|X_S)$ measures the remaining mutual information between $X_i,X_j$ and $Y$ when $X_S$ is given.
Note that unlike the bivariate mutual information, $MI(X_i;X_j;Y|X_S)$ can be negative.
When $X_S$ (each $x_S\in X_S$ containing $m$ pixels) is given, we can roughly understand the conditional mutual information $MI(X_{\{i,j\}};Y|X_S)$ as the additional benefits from $X_i$ and $X_j$ to classification.
We prove that  $MI(X_{\{i,j\}};Y|X_S)$ can be decomposed into the exclusive benefits of $X_i$ (\emph{i.e.} $MI(X_i;Y|X_j,X_S)$), the exclusive benefits of $X_j$ (\emph{i.e.} $MI(X_j;Y|X_i,X_S)$), and the benefit shared by $X_i$ and $X_j$ (\emph{i.e.} $MI(X_i;X_j;Y|X_S)$).
Thus, $MI(X_i;X_j;Y|X_S)$ can be considered as the benefits from the interaction between $X_i$ and $X_j$.
\begin{align}
MI(X_{\{i,j\}};Y|X_S)=MI(X_i;Y|X_j,X_S)+MI(X_j;Y|X_i,X_S)+MI(X_i;X_j;Y|X_S)
\end{align}

$\bullet\;$\emph{Proof}:
\begin{small}
\begin{align*}
	\text{right} &=MI(X_i;Y|X_j,X_S)+MI(X_j;Y|X_i,X_S)+MI(X_i;X_j;Y|X_S)\\
	&=MI(X_i;Y|X_j,X_S)+MI(X_j;Y|X_i,X_S)+MI(X_i;Y|X_S)-MI(X_i;Y|X_j,X_S)\\
	&= MI(X_i;Y|X_j,X_S)+MI(X_i;Y|X_S)\\
	&= \sum_{x_i,x_j,x_S,y} p(x_i,x_j,x_S,y)\log \frac{p(x_j,y|x_i,x_S)}{p(x_j|x_i,x_S)p(y|x_i,x_S)} + \sum_{x_i,x_S,y} p(x_i,x_S,y) \log \frac{p(x_i,y|x_S)}{p(x_i|x_S)p(y|x_S)}\\
	&= \sum_{x_i,x_j,x_S,y} p(x_i,x_j,x_S,y)\log \frac{p(x_j,y|x_i,x_S)}{p(x_j|x_i,x_S)p(y|x_i,x_S)} +\!\!\! \sum_{x_i,x_j,x_S,y} p(x_i,x_j, x_S,y) \log \frac{p(x_i,y|x_S)}{p(x_i|x_S)p(y|x_S)}\\
	&= \sum_{x_i,x_j,x_S,y} p(x_i,x_j,x_S,y) \log \frac{p(x_j,y|x_i,x_S)p(x_i,y|x_S)}{p(x_j|x_i,x_S)p(y|x_i,x_S)p(x_i|x_S)p(y|x_S)}\\
	&= \sum_{x_i,x_j,x_S,y} p(x_i,x_j,x_S,y) \log \frac{p(x_j,y|x_i,x_S)p(x_i,y|x_S)p(x_i,x_S)p(x_S)}{p(x_j|x_i,x_S)p(y|x_i,x_S)p(x_i|x_S)p(y|x_S)p(x_i,x_S)p(x_S)}\\
	&= \sum_{x_i,x_j,x_S,y} p(x_i,x_j,x_S,y) \log \frac{p(x_i,x_j,x_S,y)p(x_i,x_S,y)}{p(x_j|x_i,x_S)p(x_i,x_S,y)p(x_i|x_S)p(y|x_S)p(x_S)}\\
	&= \sum_{x_i,x_j,x_S,y} p(x_i,x_j,x_S,y) \log \frac{p(x_i,x_j,x_S,y)}{p(x_j|x_i,x_S)p(x_i|x_S)p(y|x_S)p(x_S)}\\
	&= \sum_{x_i,x_j,x_S,y} p(x_i,x_j,x_S,y) \log \frac{p(x_i,x_j,y|x_S)}{p(x_i,x_j|x_S)p(y|x_S)}\\
	&= MI(X_{\{i,j\}};Y|X_S) = \text{left}
\end{align*}
\end{small}
$\hfill\square$

\section{Related works about interactions}

In Section 2 of the paper, we have discussed related works about understandings of adversarial attacks, defense, and robustness.
Due to the page limit, we discuss related works about interactions in this section of the supplementary material.
Unlike previous studies about interactions, we firstly use the interaction to explain adversarial perturbations and robustness, and provide a unified view to understand existing defense methods.

Interactions between input variables of a DNN have been widely investigated in recent years.
In game theory, \citet{grabisch1999axiomatic} and \citet{lundberg2018consistent} proposed and used the Shapley interaction index based on Shapley values~\cite{shapley1953value}.
\citet{covert2020understanding} investigated the relationship between the Shapley value and the mutual information.
\citet{sorokina2008detecting} measured the interaction of multiple input variables in an additive model.
\citet{tsang2018detecting} calculated interactions of weights in a DNN.
\citet{wang2020unified} applied the interaction of adversarial perturbations to understand adversarial transferabilitiy.
\citet{murdoch2018beyond,singh2018hierarchical}, and \citet{jin2019towards} used the contextual decomposition (CD) technique to extract variable interactions.
\citet{cui2019learning} proposed a non-parametric
probabilistic method to measure interactions using a Bayesian neural network.
\citet{janizek2020explaining} extended the Integrated Gradients method~\cite{sundararajan2017axiomatic} to explain pairwise feature interactions in DNNs.
\citet{sundararajan2020shapley} defined the Shapley-Taylor index to measure interactions over binary features.
In comparison, we novelly use the multi-order interaction to understand the detailed interaction behaviors \emph{w.r.t.} adversarial attacks, which enables us to explain adversarial examples and adversarial training.

\section{Information reflected by $\Delta I^{(m)}$}
In section 4.1 of the paper, we propose the metric $I^{(m)}=\mathbb{E}_{x\in\Omega} \mathbb{E}_{i,j}[I^{(m)}_{ij}(x)]$, and $\Delta I^{(m)}\overset{\text{def}}{=}I^{(m)}_{\text{nor}}-I^{(m)}_{\text{adv}}$, which measures the difference in interactions between normal samples and adversarial examples.
In this section of the supplementary material, we prove the following property of the metric $\Delta I^{(m)}$.
\begin{align}
\Delta I^{(m)}=\mathbb{E}_{x\in\Omega}\mathbb{E}_{i,j}[\Delta I^{(m)}_{ij}(x)]
\end{align}
where $\Delta I^{(m)}_{ij}(x)=I^{(m)}_{ij}(x)-I^{(m)}_{ij}(x^{\text{adv}})$.

$\bullet\;$\emph{Proof}:
\begin{align*}
\Delta I^{(m)} &= I^{(m)}_{\text{nor}}-I^{(m)}_{\text{adv}}\\
&= \mathbb{E}_{x\in\Omega_{\text{nor}}} \mathbb{E}_{i,j}[I^{(m)}_{ij}(x)]-\mathbb{E}_{x\in\Omega_{\text{adv}}} \mathbb{E}_{i,j}[I^{(m)}_{ij}(x)]\\
&= \mathbb{E}_{x\in\Omega_{\text{nor}}} \mathbb{E}_{i,j}[I^{(m)}_{ij}(x)]-\mathbb{E}_{x\in\Omega_{\text{nor}}} \mathbb{E}_{i,j}[I^{(m)}_{ij}(x+\Delta x)]\\
&= \mathbb{E}_{x\in\Omega_{\text{nor}}} \mathbb{E}_{i,j}[I^{(m)}_{ij}(x)-I^{(m)}_{ij}(x+\Delta x)]\\
&= \mathbb{E}_{x\in\Omega_{\text{nor}}} \mathbb{E}_{i,j}[I^{(m)}_{ij}(x)-I^{(m)}_{ij}(x^{\text{adv}})]\\
&=  \mathbb{E}_{x\in\Omega_{\text{nor}}} \mathbb{E}_{i,j} [\Delta I^{(m)}_{ij}(x)]
\end{align*}$\hfill\square$

\section{Relationship between $\hat{v}(S)$ and $v(S)$}

In Section 4.1 of the main paper, we claim that the trend of $v(S)$ can roughly reflect the negative trend of $\hat{v}(S)$. In this section, we discuss the negative correlation between $\hat{v}(S)$ and $v(S)$.

According to Proposition 1, $\hat{v}(S) =H(Y|X_S)$ denotes the entropy of the classification probability given variables in $S$ of the image $x$.
Thus, $\hat{v}(S)$ measures the uncertainty of the prediction.
If the model prediction is correct and confident, \emph{i.e.} the value of $v(S)=\log p(y=y^\text{truth}|x,S)$ is large, then the uncertainty $\hat{v}(S)$ is very low.
In comparison, if the model prediction is correct but with a small value of $v(S)$, then the uncertainty is large, yielding a large value of $\hat{v}(S)$.
Therefore, the trend of $v(S)$ can roughly reflect the negative trend of $\hat{v}(S)$ when the model prediction is correct.

\section{More discussions about the disentanglement metric}
This section provides more discussions about the disentanglement metric in Eq. (4) of the main paper.
The motivation of the disentanglement is to measure the discrimination power of interactions of a specific order, as discussed in Section 4.2 of the paper.
According to the efficiency property of the multi-order interactions, the model output can be decomposed into the weighted sum of massive interaction components $\Delta v(i,j,S|x)$:

\begin{equation}
\begin{small}
v(N|x)\! =v(\emptyset|x)+\sum_{i\in N}\phi^{(0)}(i|x)+\!\!\sum_{i,j\in N, i\ne j}\sum_{m=0}^{n-2}\frac{n-1-m}{n(n-1)}\mathbb{E}_{S\subseteq N\setminus\{i,j\},|S|=m}[\Delta v(i,j,S)]
\end{small}
\end{equation}
In this way, if interaction components of a certain order $m$ are all positive (or negative) and do not conflict with each other, it indicates that these components jointly promote or suppress the model output, showing a strong discrimination power. Otherwise, if some interaction components are positive and others are negative, their effects on the model output will be eliminated. In this case, the discrimination power of interactions is poor.

Therefore, we design the disentanglement metric $D^{(m)}$ in Eq. (4) to model the above phenomenon. The physical meaning of this metric is shown in the following equation.
\begin{align}
D^{(m)}=\mathbb{E}_{x\in \Omega}\mathbb{E}_{i,j\in N, i\ne j} \frac{\overbrace{|\mathbb{E}_{S\subseteq N\setminus\{i,j\}, |S|=m} \Delta v(i,j,S|x)|}^{\text{the strength of the average utility of all components}}}{\mathbb{E}_{S\subseteq N\setminus\{i,j\}, |S|=m} \underbrace{|\Delta v(i,j,S|x)|}_{\text{the strength of each component}}}
\end{align}
The numerator measures the strength of the average utility of all interaction components between $(i,j)$ under different contexts $S$. The dominator represents the average strength of each interaction component. If $D^{(m)}$  approximates to 1, then it indicates that almost all interaction components have similar effects (either positive or negative) on the model output. If $D^{(m)}$ approximates to 0, then it shows that most interaction components conflict with each other and are eliminated. Therefore, the disentanglement $D^{(m)}$  measures the discrimination power of interactions.

\section{The attribution-based method of detecting adversarial examples}

In Section 4.3 of the paper, we claim that the attribution score used in~\cite{DBLP:journals/corr/abs-1906-03499} to detect adversarial exmaples can be writtern as $\phi^{(n-1)}(i|x)$.
In the supplementary material, this section provides proofs for this claim.

\citet{DBLP:journals/corr/abs-1906-03499} proposed an attribution-based method to detect adversarial examples, which used the attribution score of input variables.
The attribution score of the variable $i$ in \cite{DBLP:journals/corr/abs-1906-03499} is defined as
\begin{align}
\phi(x)_i := f(x)_c- f(x_{(i)})_c,~~\text{where}~~ c=\arg \max_{j\in C} f(x)_j
\end{align}
where $x$ denotes the original input sample, and $x_{(i)}$ denotes the input sample with the $i$-th variable masked by 0.
$f(x)_c$ denotes the network output of the $c$-th category.
Actually, $f(x)_c$ can also be written as $v(N|x)$, and $f(x_{(i)})_c$ can be written as $v(N\setminus\{i\}|x)$.
Thus, the attribution score can be represented as $v(N|x)-v(N\setminus\{i\}|x)$.
We prove that $v(N|x)-v(N\setminus\{i\}|x)=\phi^{(n-1)}(i|x)$.

$\bullet\;$\emph{Proof}:
\begin{align*}
v(N|x)-v(N\setminus\{i\}|x) &= v((N\setminus\{i\})\cup \{i\}|x)-v(N\setminus\{i\}|x)\\
&= v(S\cup \{i\}|x)-v(S|x) \qquad\%~ S\overset{\text{def}}{=} N\setminus\{i\}, |S|=n-1\\
&= \mathbb{E}_{S\subseteq N\setminus\{i\},|S|=n-1} \left[v(S\cup \{i\}|x)-v(S|x)\right]\\
&= \phi^{(n-1)}(i|x) 	
\end{align*}$\hfill\square$

According to the \textit{accumulation property} of multi-order Shapley values and interactions, we have $\phi^{(n-1)}(i|x)=\mathbb{E}_{j\in N\setminus\{i\}} \left[\sum_{m=0}^{n-2} I^{(m)}_{ij}\right]+\phi^{(0)}(i|x)$.
This indicates that $\phi^{(n-1)}(i|x)$ contains the interaction components with the highest order $(m=n-2)$, which are not included in Shapley values with orders lower than $n-1$.
Section 4.1 of the paper has pointed that high-order interactions are the most sensitive to adversarial perturbations, thereby enabling the detection of adversarial examples.

\section{Effectiveness of the dropout method to alleviate adversarial utilities}
In Section 4.3 of the paper, we claim that the dropout operation mainly hurts high-order interactions, rather than low-order interactions.
In this section, we theoretically prove such effects of  the dropout operation.

Given the input sample $x\in \mathbb{R}^n$ and the dropout rate $\alpha$, let $\mathcal{K}=\{K|K\subset N,|K|=\lfloor(1-\alpha)n\rfloor\}$ denote all possible sets of remained variables after the dropout operation.
Let $v^\alpha(N|x)=\mathbb{E}_{K\in\mathcal{K}} [v(K|x)]$ denote the average network output among all inputs after the dropout operation with rate $\alpha$.
According to the efficiency property of the multi-order interaction, we have
\begin{equation}
v(N|x) = v(\emptyset|x)+ \sum_{i\in N} \phi^{(0)} (i|x)+\sum_{i\ne j \in N} \sum_{m=0}^{n-2} \frac{n-1-m}{n(n-1)}I^{(m)}_{ij,N}(x)
\end{equation}
where $I^{(m)}_{ij,N}(x)$ denotes the $m$-order interaction between variables $(i,j)$ of the input $x$ with all variables $N$.
Similarly,
\begin{equation}
v(K|x) = v(\emptyset|x)+\sum_{i\in K} \phi^{(0)} (i|K,x)+\sum_{i\ne j \in K} \sum_{m=0}^{k-2} \frac{k-1-m}{k(k-1)}I^{(m)}_{ij,K}(x)
\end{equation}
where $k=|K|=\lfloor (1-\alpha)n\rfloor$.
Thus,
\begin{equation}
\begin{small}
	\begin{aligned}
		v^\alpha(N|x) &= \mathbb{E}_{K\in\mathcal{K}}[v(K|x)]\\
		&=\mathbb{E}_{K\in\mathcal{K}}\left[v(\emptyset|x)+\sum_{i\in K} \phi^{(0)}(i|K,x) +\sum_{i\ne j \in K}\sum_{m=0}^{k-2} \frac{k-1-m}{k(k-1)}I^{(m)}_{ij,K}(x)\right]\\
		&=v(\emptyset|x)+\mathbb{E}_{K\in\mathcal{K}}\left[\sum_{i\in K} \left(v(i|x)-v(\emptyset|x)\right)\right]+\mathbb{E}_{K\in\mathcal{K}}\left[\sum_{i\ne j \in K}\sum_{m=0}^{k-2}\frac{k-1-m}{k(k-1)}I^{(m)}_{ij,K}(x)\right]\\
		&=v(\emptyset|x)+\mathbb{E}_{\substack{K\subset N\\k=(1-\alpha)n}}\left[\sum_{i\in K} \phi^{(0)}(i|x)\right]+\mathbb{E}_{\substack{K\subset N\\k=(1-\alpha)n}}\left[\sum_{i\ne j \in K}\sum_{m=0}^{k-2}\frac{k-1-m}{k(k-1)}I^{(m)}_{ij,K}(x)\right]\\
		&=v(\emptyset|x)+(1-\alpha)\sum_{i\in N} \phi^{(0)}(i|x)+\mathbb{E}_{\substack{K\subset N\\k=(1-\alpha)n}}\left[\sum_{i\ne j \in K}\sum_{m=0}^{k-2}\frac{k-1-m}{k(k-1)}I^{(m)}_{ij,K}(x)\right]\\
		&=v(\emptyset|x)+(1-\alpha)\sum_{i\in N} \phi^{(0)}(i|x)+\mathbb{E}_{\substack{K\subset N\\k=(1-\alpha)n}}\left[\sum_{i\ne j \in K}\sum_{m=0}^{k-2}\frac{k-1-m}{k(k-1)}\mathbb{E}_{\substack{S\subseteq K\setminus\{i,j\}\\|S|=m}}\left[\Delta v(i,j,S|x)\right]\right]\\
		&=v(\emptyset|x)+(1-\alpha)\sum_{i\in N} \phi^{(0)}(i|x)+\frac{k(k-1)}{n(n-1)}\sum_{i\ne j \in N}\sum_{m=0}^{k-2}\frac{k-1-m}{k(k-1)}\mathbb{E}_{\substack{K\subset N\\k=(1-\alpha)n}}\mathbb{E}_{\substack{S\subseteq K\setminus\{i,j\}\\|S|=m}}\left[\Delta v(i,j,S|x)\right]\\
		&=v(\emptyset|x)+(1-\alpha)\sum_{i\in N} \phi^{(0)}(i|x)+\sum_{i\ne j\in N} \sum_{m=0}^{k-2} \frac{k-1-m}{n(n-1)}\mathbb{E}_{\substack{K\subset N\\k=(1-\alpha)n}}\mathbb{E}_{\substack{S\subseteq K\setminus\{i,j\}\\|S|=m}}\left[\Delta v(i,j,S|x)\right]\\
		&=v(\emptyset|x)+(1-\alpha)\sum_{i\in N} \phi^{(0)}(i|x)+\sum_{i\ne j\in N} \sum_{m=0}^{k-2} \frac{k-1-m}{n(n-1)} \mathbb{E}_{\substack{S \subset N\setminus \{i,j\}\\|S|=m}} \left[\Delta v(i,j,S|x)\right]\\
		&=v(\emptyset|x)+(1-\alpha)\sum_{i\in N} \phi^{(0)}(i|x)+\sum_{i\ne j\in N} \sum_{m=0}^{k-2} \frac{k-1-m}{n(n-1)} I^{(m)}_{ij,N}(x)
	\end{aligned}
\end{small}		
\end{equation}
Thus, the change in the network output caused by the dropout operation can be represented as follows,
\begin{equation}
\begin{small}
	\begin{aligned}
		v(N|x)\!-\! v^\alpha(N|x) &= \alpha \sum_{i\in N}\phi^{(0)}(i,x)+\!\!\sum_{i\ne j \in N} \left[\sum_{m=0}^{n-2} \frac{n-1-m}{n(n-1)}I^{(m)}_{ij,N}(x) - \sum_{m=0}^{k-2} \frac{k-1-m}{n(n-1)} I^{(m)}_{ij,N}(x)\right]\\
		&= \alpha \sum_{i\in N}\phi^{(0)}(i,x) +\!\!\sum_{i\ne j\in N}\sum_{m=0}^{k-2}\frac{n-k}{n(n-1)} I^{(m)}_{ij,N}(x) + \sum_{i\ne j \in N}\sum_{m=k-1}^{n-2}\frac{n-1-m}{n(n-1)}I^{(m)}_{ij,N}(x)\\
		&=\alpha \sum_{i\in N}\phi^{(0)}(i,x)+\!\! \frac{\alpha}{n-1}\sum_{i\ne j \in N}\sum_{m=0}^{k-2} I^{(m)}_{ij,N}(x) +\!\! \underbrace{\sum_{i\ne j \in N}\sum_{m=k-1}^{n-2}\frac{n-1-m}{n(n-1)}I^{(m)}_{ij,N}(x)}_{\text{ high-order interactions}}\\
	\end{aligned}
\end{small}
\label{eq:dropout-I}
\end{equation}
According to Eq.~\eqref{eq:dropout-I}, the dropout operation removes all high-order interactions ($m \! > \! (1-\alpha)n \! -\ \! 2$), while slightly affects low-order interactions.
Thus, the dropout operation can remove sensitive interaction components of the DNN, thereby reducing the attacking utility of perturbations and correcting the network output.

\section{More experimental results}
\label{sec:exp_results}

\subsection{More visualization results based on ResNet-50}

The method in~\cite{zhang2020interpreting-dropout} could be directly extended to visualize salient concepts of the multi-order interaction without much change.
For a specific order $m$, we visualized all contexts $\{S\}$ of $I^{(m)}(i,j)$ for each pair $(i,j)$ with top $10\%$ interaction strengths. We exclusively visualized contexts $\{S\}$ that boosted the strength of $I^{(m)}(i,j)$,~\emph{i.e.} $ \Delta v(S,i,j)\cdot I^{(m)}(i,j)>0$. Let $\textrm{map}(S)\in\{0,1\}^{n}$ denote the map corresponding to the context $S$. If the $k$-th pixel was contained in $S$, then $\textrm{map}_{k}(S)=1$; otherwise, $\textrm{map}_{k}(S)=0$. In this way, we visualized the weighted average contexts \emph{w.r.t.} pixels $(i,j)$ as $\sum_{S \subseteq N,|S|=m}|\Delta v(S,i,j)| \cdot \textrm{map}(S)$.

\begin{figure}[t]
\begin{minipage}{0.63\linewidth}
	\centering
	\includegraphics[width=\linewidth]{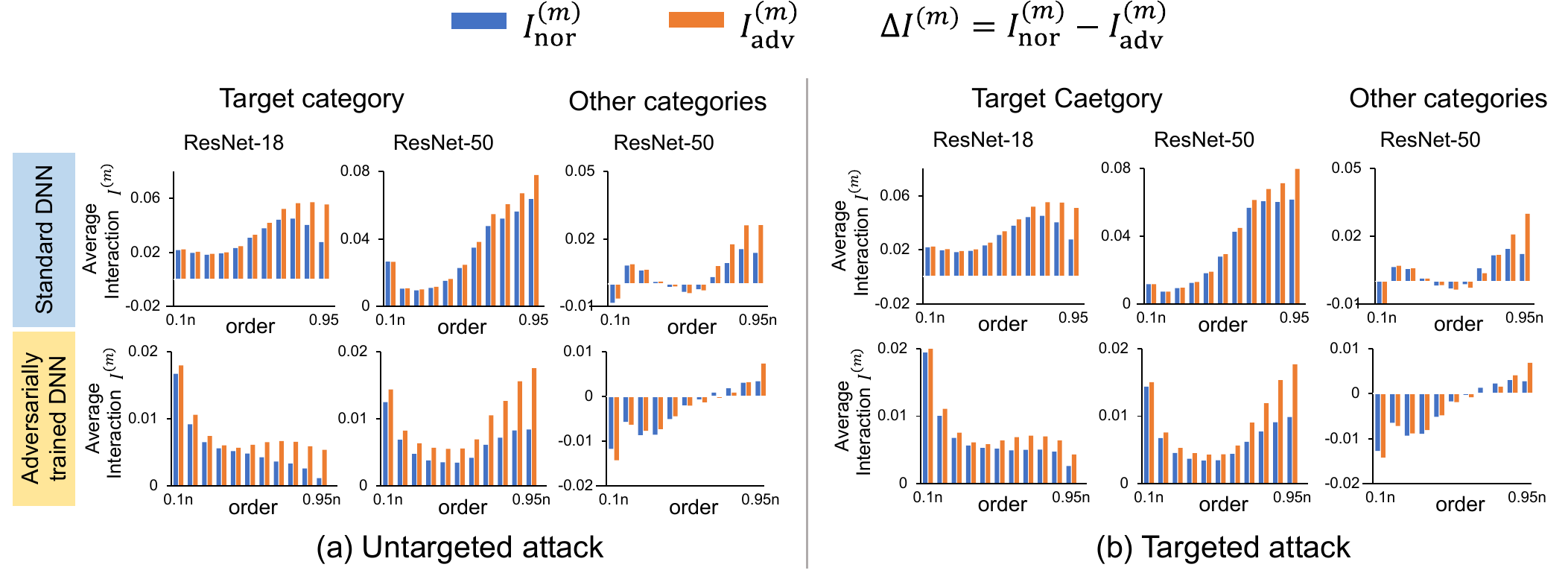}
	\vspace{-20pt}
	\caption{Interactions \emph{w.r.t.} the target category and other categories.}
	\label{fig:other_category}
\end{minipage}
\hfill
\begin{minipage}{0.33\linewidth}
	\centering
	\includegraphics[width=\linewidth]{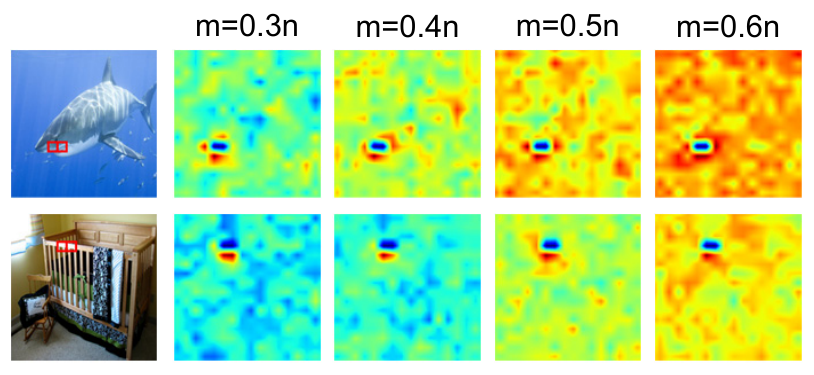}
	\vspace{-10pt}
	\caption{Contexts of the m-order interaction $I^{(m)}_{ij}$ in normal samples of standard ResNet-50.}
	\vspace{-5pt}
	\label{fig:vis_res50}
\end{minipage}
\end{figure}

Figure 2 (right) in the paper has shown visualization results on ResNet-18.
Here, we presented more visualization results of interaction contexts based on ResNet-50~\cite{he2016deep} in
Figure~\ref{fig:vis_res50}.
As Figure~\ref{fig:vis_res50} shows, low-order interactions usually represented simple features of local collaborations, and high-order interactions usually reflected complex features of global collaborations.

\subsection{More discussions about the difference between standard DNNs and adversarially trained DNNs in Figure 3 of the main paper}

In Figure 3 of the main paper, it seems that standard DNNs and adversarially trained DNNs are similar under the metric $\Delta I^{(m)}$, but Figure 3 does reflect the significant difference of interaction behaviors between standard DNNs and adversarially trained DNNs.
Figure 3 shows that although high-order interactions are penalized in both standard and adversarially trained DNNs, high-order interactions in the standard DNN decreased much more than those in the adversarially trained DNN.

More specifically, we define the metric  $\frac{|\Delta I^{(m)}|}{\sum_{m^\prime} |\Delta I^{(m^\prime)}|}$ to compare the change of multi-order interactions in different DNNs, where $\Delta I^{(m)}=I_{\text{ori}}^{(m)}-I_{\text{adv}}^{(m)}$ measures the difference of the $m$-th order interaction between normal samples and adversarial examples. $\sum_{m^\prime} |\Delta I^{(m^\prime)}|$ is the total strength of differences of multi-order interactions, which is used for normalization.
Table~\ref{tab:normed} shows that the interaction behaviors in standard DNNs and adversarially trained DNNs were quite different. In the standard ResNet-18, low-order interactions were penalized slightly, and high-order interactions dropped a lot. In comparison, in the adversarially trained ResNet-18, low-order interactions and middle-order interactions were penalized more than those in the standard ResNet-18, while high-order interactions were less penalized than those in the standard ResNet-18. Therefore, the adversarially trained DNN was supposed to exhibit a larger value of $\frac{|\Delta I^{(m)}|}{\sum_{m^\prime} |\Delta I^{(m^\prime)}|}$ for low-order interactions (with small $m$) than the standard DNN. In comparison, the adversarially trained DNN exhibited a smaller value of $\frac{|\Delta I^{(m)}|}{\sum_{m^\prime} |\Delta I^{(m^\prime)}|}$ for high-order interactions (with large $m$) than the standard DNN.

\begin{figure}[t]
\centering
\begin{minipage}{\linewidth}
    \captionof{table}{Interaction behaviors in standard DNNs and adversarially trained DNNs are quite different \emph{w.r.t.} the metric $\frac{|\Delta I^{(m)}|}{\sum_{m^\prime} |\Delta I^{(m^\prime)}|}$.
    \vspace{-5pt}}
    \label{tab:normed}
    \resizebox{\linewidth}{!}{
    \begin{tabular}{c|c c c c c c c c c c c}
    \hline
        $m$ & $0.1n$ & $0.2n$ & $0.3n$ & $0.4n$ & $0.5n$ & $0.6n$ & $0.7n$ & $0.8n$ & $0.85n$ & $0.9n$ & $0.95n$\\
        \hline
        $\frac{|\Delta I^{(m)}|}{\sum_{m^\prime} |\Delta I^{(m^\prime)}|}$  & \multirow{3}{*}{0.003} & \multirow{3}{*}{0.005} & \multirow{3}{*}{0.008} & \multirow{3}{*}{0.014} &\multirow{3}{*}{0.021} & \multirow{3}{*}{0.040} & \multirow{3}{*}{0.066} & \multirow{3}{*}{0.103} & \multirow{3}{*}{\textbf{0.138}} &  \multirow{3}{*}{\textbf{0.219}} & \multirow{3}{*}{\textbf{0.383}}\\
        in standard & {} & {} & {} & {} & {} & {} & {} & {} & {} & {} & {} \\
        ResNet-18 & {} & {} & {} & {} & {} & {} & {} & {} & {} & {} & {} \\
        \hline
        $\frac{|\Delta I^{(m)}|}{\sum_{m^\prime} |\Delta I^{(m^\prime)}|}$ in & \multirow{3}{*}{\textbf{0.016}} & \multirow{3}{*}{\textbf{0.023}} & \multirow{3}{*}{\textbf{0.055}} & \multirow{3}{*}{\textbf{0.082}} & \multirow{3}{*}{\textbf{0.095}} & \multirow{3}{*}{\textbf{0.100}} & \multirow{3}{*}{\textbf{0.100}} & \multirow{3}{*}{\textbf{0.108}} & \multirow{3}{*}{0.117} & \multirow{3}{*}{0.144} & \multirow{3}{*}{0.160}\\
        adversarially & {} & {} & {} & {} & {} & {} & {} & {} & {} & {} & {} \\
        trained ResNet-18 & {} & {} & {} & {} & {} & {} & {} & {} & {} & {} & {} \\
        \hline
    \end{tabular}
    }
\end{minipage}
\end{figure}

\subsection{Discussions about attacking utilities of multi-order interactions.}

Figure 4 of the paper shows that low-order interactions in adversarially trained DNNs usually have more attacking utilities.
Note that the attacking utility on high/middle-order interactions may make $\Delta J^{(m)}$ on low-order interactions negative, as a trade-off.
According to Eq. (3), the sum of attacking utilities on low-order, middle-order, and high-order interactions is a constant ($\Delta v(N|x)$).
Therefore, the penalization of high-order interactions may also cause the increase of low-order interactions to some extent. In this way, there are two effects on low-order interactions. First, adversarial attacks penalize low-order interactions. Second, the penalization of high-order interactions also boosts low-order interactions as a side effect. In such a trade-off, low-order interactions decrease in most DNNs. In this way, low-order interactions are increased in a few special cases, which can be explained by the above analysis.

\subsection{More details about the approximation method of computing interactions}

To reduce the computational cost, we applied the sampling-based approximation method in \cite{zhang2020interpreting-dropout} and did not compute interactions at the pixel-wise level. Instead, we split the image into $16\times 16$ grids, and took each grid as a single input variable, thereby $n=256$.
Then, we randomly sampled 200 pairs of grids $(i,j)$, and for each pair of grids and each order $m$, we sampled contexts $S$ s.t. $|S|=m$ for 100 times to approximate the interaction.
Then we computed the $m$-th order interactions based on Eq. (2) in the main paper by using the sampled pairs of grids and contexts.

\subsection{More details about the setting of masking in the computation of interactions}

\begin{figure}[t]
\centering
\includegraphics[width=0.9\linewidth]{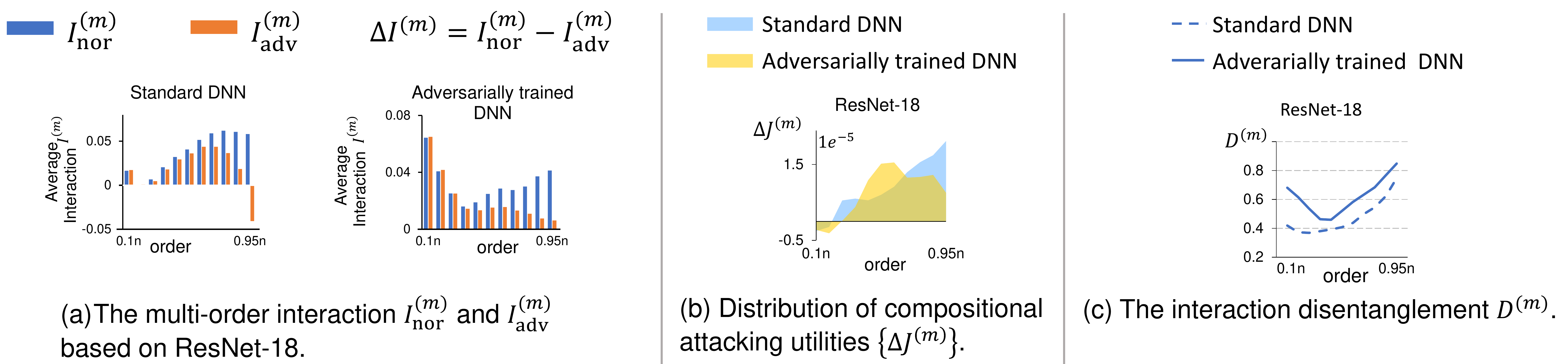}
\vspace{-5pt}
\caption{Experimental results on ResNet-18 when setting the masked variables to zero.}
\vspace{-5pt}
\label{fig:zero}
\end{figure}

In the main paper, we measured the interaction  $I^{(m)}$ by setting $v(S|x)=\log p(y=y^\text{truth}|\text{given variables in}~S~\text{in the input}~x~\text{and mask variables in}~N\setminus S)$.
Variables not in $S$ were set to the average value over different input samples following settings in \cite{ancona2019explaining}, to represent their absence.
In this section, we conducted experiments by using setting masked variables to zero, to verify that the effects of the choice of masking on the results did not affect our conclusions.
Figure~\ref{fig:zero} shows that the choice of masking did not affect the conclusion that adversarial attacks mainly affect high-order interactions.

\subsection{Analysis about interactions \emph{w.r.t.} the target category and other categories}

In the main paper, experimental results on the ground-truth category have verified our conclusions that adversarial attacks mainly affect high-order interactions.
In this section of the supplementary material, we show more results on the target category and other categories, which also verified our conclusions.

In the targeted attack, the target category was referred to as the target label $y^{\text{target}}\ne y^{\text{truth}}$.
In the untargeted attack, we considered the misclassified category $y^{\text{pred}}\ne y^{\text{truth}}$ as the target category.
Given each input image, we measured the interaction \emph{w.r.t.} the output of its target category by setting $v(S|x)=\log p(y=y^\text{target}|S,x)$.
Besides, we also measured the interactions \emph{w.r.t.} outputs of other categories, except the ground-truth category.
Considering the softmax operation $p(y=y^\text{truth}|S,x)= \frac{\exp(h_{y^\text{truth}}(S|x))}{\sum_{y^\prime} \exp(h_{y^\prime}(S|x))}$,
we set $v(S|x)=\log \sum_{y^\prime \ne y^{\text{truth}}} \exp(h_{y^\prime}(S|x))$ to measure the interaction \emph{w.r.t.} effects on other categories.
$h_{y^\prime}(S|x)$ denoted the network output of the catagory $y^\prime$ before the softmax layer, when we took variables in $S$ of $x$ as the input.
Figure~\ref{fig:other_category} shows interactions \emph{w.r.t.} outputs of the target category and other categories.
We found that high-order interactions \emph{w.r.t.} other categories usually increased.
Meanwhile, high-order interactions \emph{w.r.t.} the target category also significantly increased.
This indicated that adversarial perturbations adversely affected complex features corresponding to the ground-truth category, while encouraging features for other categories, especially for the target category.

\subsection{Extended experiments on the relationship between the strength of attacks and the change of high-order interactions}
In Section 4.1 of the paper, we find that adversarial attacks mainly affect high-order interactions in DNNs.
In this section, we further explore the relationship between the strength of attacks and the change of high-order interactions.

Using the standard ResNet-18 learned on the ImageNet dataset, we conducted untargeted PGD attacks on input samples with different attacking strengths, in order to test the effects of attacking strength on high-order interactions. The strength of attacks was represented by the iteration numbers of the PGD attack. Then, we computed the change in high-order interactions $\Delta I^{(m)}$ caused by adversarial attacks of different strengths. We also computed the Pearson correlation coefficient between the strength of attacks and  $\Delta I^{(m)}$.
Table~\ref{tab:strength-change} shows a close relationship between changes in high-order interactions and the strength of adversarial attacks.
Besides, for both weak and strong attacks, the highest-order (here we set it as $0.95n$-order) interactions were much more sensitive than interactions of not-so-high orders (e.g., $0.7n$-order).

\begin{table}[t]
\centering
\caption{The close relationship between the attacking strength and changes of high-order interactions.}
\label{tab:strength-change}
\resizebox{0.9\linewidth}{!}{
\begin{tabular}{c|c c c c c}
    \hline
    Attacking strength (iteration number) &  $\Delta I^{(0.7n)}$ &  $\Delta I^{(0.8n)}$ &  $\Delta I^{(0.85n)}$ &  $\Delta I^{(0.9n)}$ &  $\Delta I^{(0.95n)}$\\
    \hline
    4 & 0.0518 & 0.0990 & 0.1411 & 0.1944 & \textbf{0.2274}\\
    8 & 0.0789 & 0.1741 & 0.2399 & 0.3118 & \textbf{0.3632}\\
    16 & 0.1113 & 0.2565 & 0.3342 & 0.4157 & \textbf{0.4657}\\
    32 & 0.1512 & 0.3338 & 0.4391 & 0.5480& \textbf{0.6085}\\
    Pearson correlation coefficient & 0.9799 & 0.9603 & 0.9635 & 0.9667& 0.9634\\
    \hline
\end{tabular}
}
\end{table}

\subsection{Extended experiments based on models with certified robustness}

In the main paper, experimental results on DNNs trained via adversarial training~\cite{madry2018towards} have verified our conclusions that adversarial attacks mainly affect high-order interactions, and low-order interactions in adversarial examples towards robust DNNs have more attacking utilities.
In this section, we conducted extended experiments on other defending method~\cite{cohen2019certified}.
Given the ResNet-50, which was trained using randomized smoothing~\cite{cohen2019certified}, we generated adversarial examples using the PGD attack on samples from the validation set of the ImageNet dataset. We followed experimental settings of the classic PGD attack (see Section 4.1 in the main paper). Then, we computed multi-order interactions in normal samples and adversarial examples on the pre-trained ResNet-50.

\begin{figure}[t]
\centering
\includegraphics[width=0.7\linewidth]{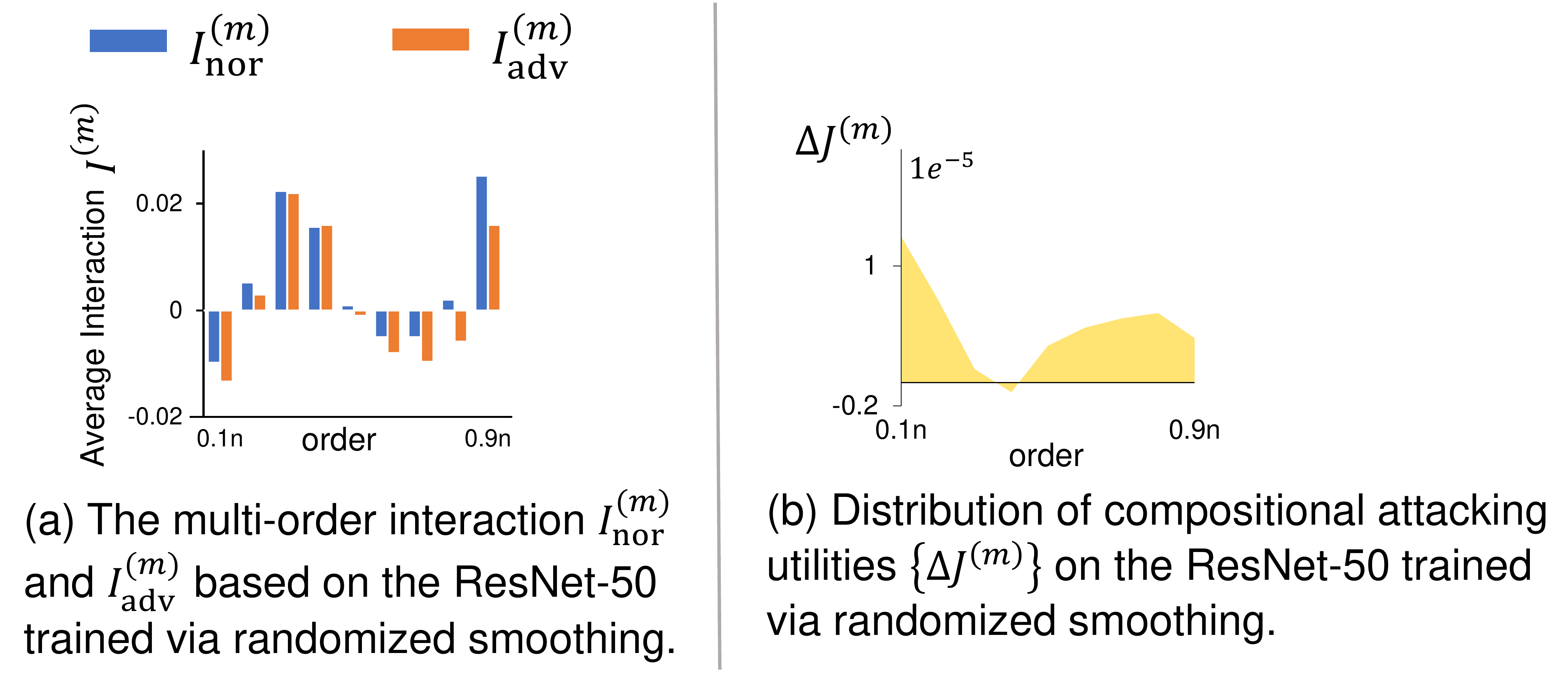}
\vspace{-5pt}
\caption{Experimental results on the ResNet-50 trained via randomized smoothing.}
\vspace{-5pt}
\label{fig:random-smoothing}
\end{figure}

Figure~\ref{fig:random-smoothing} shows that in the model trained via randomized smoothing, adversarial attacks also mainly affected high-order interactions, and low-order interactions had more attacking utilities on the model output, which verified our conclusions.

\subsection{More discussions about the comparison with frequency-based methods}

In Section 4.1 of the main paper, we have compared our interaction-based metric with the frequency-based metric. This section provides more discussions about the comparative experiment.

Previous studies~\cite{yin2019fourier,wang2020high,harder2021spectraldefense} explained adversarial perturbations as high-frequency features.
In this paper, we compared the frequency metric with the interaction metric, and showed that our interaction metric could better explain the essential property of adversarial perturbations.
In order to measure the significance of features of different frequencies in an input image $x$, we applied the Fourier transform to the image to obtain the frequency spectrum $F$.
Then, we shifted the low frequency components to the center of the spectrum.
The magnitude of the $r$-frequency component was quantified as $F^{(r)} = \sqrt{\sum_{(h,w):h^2+w^2=r^2}\Vert F^{(h,w)} \Vert_2^2}$.
In order to fairly compare these two metrics, we computed the Fisher's discriminant ratio~\cite{fisher1936use} using two metrics, which measured the ratio of the variance between the classes (between normal samples and adversarial examples) to the variance within the classes.
Such experimental settings has also been introduced in Figure 6 (left) of the main paper.

\subsection{Explaining high recoverability of adversarial examples on adversarially trained DNNs}

In Section 4.3 of the paper, we claim that adversarial examples towards adversarially trained DNNs usually exhibit higher recoverability than adversarial examples
towards standard DNNs.
In the supplementary material, this section shows experimental results to verify the above claim.

We generated adversarial examples for normal validation samples in the ImageNet dataset by following settings of the untargeted PGD attack~\cite{madry2018towards}, in which $\epsilon = 16/255$, and the attack was conducted with 10 steps with the step size $2/255$.
Adversarial examples were generated  based on ResNet-18/50 and DenseNet-161 trained on the ImageNet dataset.
Then, we used same parameters to conduct the targeted PGD attack and recover normal samples.

\begin{table}[h]
\caption{The distance between normal samples $x$ and adversarial examples $x^{\text{adv}}$, and the distance between $x$ and the recovered samples $\hat{x}$.
	\label{tab:recover}}
\centering
\resizebox{0.6\linewidth}{!}{
	\begin{tabular}{c|c|c||c|c}
		\hline
		{} & \multicolumn{2}{c||}{Standard DNN} & \multicolumn{2}{c}{Adversarially trained DNN} \\
		\cline{2-5}
		{} & $\! \mathbb{E}\Vert x-x^{\text{adv}}\Vert_2 \!$ & $\!\mathbb{E}\Vert x-\hat{x}\Vert_2\!$ &  $\!\mathbb{E}\Vert x-x^{\text{adv}}\Vert_2\!$ &$ \!\mathbb{E}\Vert x-\hat{x}\Vert_2\!$ \\
		\hline
		ResNet-18 & \textbf{9.72} & 13.57 & 18.69 & \textbf{11.45}\\
		ResNet-50 & \textbf{9.56} & 13.40 & 18.34 & \textbf{12.68} \\
		DenseNet-161 & \textbf{9.67} & 13.51 & 18.55 & \textbf{13.26} \\
		\hline
	\end{tabular}
}
\end{table}

Table~\ref{tab:recover} shows that adversarially trained DNNs usually exhibited higher recoverability than standard DNNs.
This can be explained Proposition 1 in the paper.
As we have discussed in Section 4.1 of the paper, adversarial perturbations towards adversarially trained DNNs usually pay more attention to low-order interactions than perturbations towards standard DNNs.
On the other hand, the low-order interaction $I^{(m)}_{ij}(x)$ is equivalent to the conditional mutual information $MI(X_i;X_j;Y|X_S)$ given small contexts $X_S$ according to Proposition 1, while the high-order interaction corresponds to such a mutual information conditioned on large contexts with massive variables. In general, compared to high-order interactions, low-order interactions are conditioned on less contextual variables, thereby suffering less from adversarial perturbations, \emph{i.e.} obviously $MI(X_i;X_j;Y|X_S)$ usually suffers less from adversarial perturbations, if the condition
$x_S$ only contains very few variables. In other words, low-order interactions are more transferable among different contexts $x_S$, so it is easy to invert adversarial perturbations of low-order interactions.
In this way, because adversarial perturbations for adversarially trained DNNs mainly focus on low-order interactions, such perturbations are easy to be recovered.

\end{document}